\documentclass[sigconf, nonacm]{acmart}
\usepackage[utf8]{inputenc}
\usepackage[T1,T2A]{fontenc}   
\usepackage{subcaption}        
\usepackage{algorithm2e}       
\usepackage{float}
\usepackage{graphicx}
\usepackage{caption}           

\renewcommand\footnotetextcopyrightpermission[1]{}
\acmConference[AITC 2026]{AI Transparency Conference}{June 5--6, 2026}{TBD}

\begin{document}

\title{Process then Retrieve: Investigating the Origins of the Modality Gap in VLMs}


\author{Arpita Shanbhag*}
\affiliation{%
  \institution{Algoverse}
  \city{Dublin, CA}
  \country{USA}
}
\email{arpitashanbhag01@gmail.com}

\author{Julia Tran*}
\affiliation{%
  \institution{Algoverse}
  \city{Dallas, TX}
  \country{USA}
}
\email{tranbjulia@gmail.com}

\author{Dhruv Mandala}
\affiliation{%
  \institution{Algoverse}
  \city{Fremont, CA}
  \country{USA}
}
\email{dhruvreddymandala@gmail.com}

\author{Ayda Sultan}
\affiliation{%
  \institution{Algoverse}
  \city{Addis Ababa}
  \country{Ethiopia}
}
\email{ayda@algoverseairesearch.org}



\begin{abstract}

Understanding the origins of the modality gap in vision–language models remains a significant challenge. This work investigates whether differences in information density between modality-specific tokens at the embedding layer contribute to this imbalance. We show that text tokens enter the model with much lower semantic density than visual tokens, driving early layers to focus on processing text representations before combining them with vision. Through mechanistic analysis, we demonstrate that replacing vision token positions with their semantically closest text-token counterparts elicits increased attention to those positions in early layers. These findings provide a foundation for addressing the modality gap and provide insight to improve multimodal reasoning in VLM architectures.

\end{abstract}


\maketitle

\def\thefootnote{*}\footnotetext{indicates equal contribution.}\def\thefootnote{\arabic{footnote}}


\section{Introduction}
Vision–language models (VLMs) such as \textsc{PaliGemma} \citep{beyer2024paligemma} and \textsc{Qwen}\citep{bai2023qwen} integrate large language models with pre-trained vision encoders via projection adapters to achieve strong multimodal performance, but the internal mechanisms that reveal how visual information enters and shapes the linguistic latent space remain a "black box" \citep{shukor2024}. The "modality gap," characterized by geometric separation \citep{shukor2024}, late-stage processing \citep{neo2024, venhoff2025too, nikankin2025same}, and text dominance \citep{wang2024picture} causes models to perform worse on vision-centric tasks.

Focusing on adapter-based VLMs, we show that differences in semantic density between vision and text tokens at the embedding layer lead early layers to prioritize building textual context while deferring visual information to later stages.

While recent work by \citet{venhoff2025too} identifies late-stage integration as a general phenomenon in VLMs, our work moves beyond observation to uncover the mechanisms underlying this behavior in adapter-based architectures. We first find a strong asymmetry at the embedding layer, where vision tokens are substantially more semantically dense than text tokens. We further observe that visual representations remain largely invariant across layers, whereas textual representations undergo continuous transformations. To probe the effect of this imbalance, we replace vision token embeddings with their nearest semantically similar text embeddings and find that this increases early-layer attention to the replaced position, indicating that initial differences in semantic density influence how the model allocates cross-modal attention.

\section{Related Works}
\textbf{Adapter-based VLMs}: An adapter-based VLM is structurally defined by three main components: a pre-trained vision encoder that processes the input image into discrete patch tokens, a pre-trained large language model (LLM), and a linear or MLP adapter. The adapter projects high-dimensional visual representations into the language decoder's textual latent space. While these architectures use image tokens as text to leverage the LLM's language abilities, the internal process of how visual information shapes the model remains a "black box" \citep{beyer2024paligemma}.

\textbf{The Modality Gap}: Prior work identifies a persistent "modality gap" that limits these architectures, stemming from geometric separation, late-stage processing, and text dominance. In the joint latent space, image and text modalities occupy significantly different regions, often residing in distinct narrow cones. Visual features projected by the adapter initially occupy a subspace that is nearly orthogonal to the LLM's pretrained textual embedding space \citep{shukor2024}. The gap emerges at random initialization and is sustained by an information imbalance between the modalities since images contain more detail than their corresponding text captions \citep{schrodi2024two}.

A fundamental consequence of this structural misalignment is the delayed processing of visual information, where cross-modal alignment emerges gradually and peaks in the middle-to-late layers of the network \citep{venhoff2025too}. This delayed integration manifests as a "two-hop problem" during factual reasoning tasks: visual representations are resolved too late in the computation, bypassing the early layers responsible for factual recall \citep{nikankin2025same}. Consequently, VLMs frequently exhibit text dominance, relying heavily on linguistic priors while neglecting visual evidence, causing models to perform worse on vision-centric tasks \citep{wang2024picture}.

\textbf{Mechanistic Interpretability of VLMs}: Recent interpretability research has worked to bridge this gap by probing internal pathways, revealing that models process modalities via different internal data circuits \citep{nikankin2025same}. Mechanistic interventions, such as back-patching strong visual activations into earlier layers of the text circuit or targeting semantic hub layers where cross-modal alignment peaks, have been proposed to force earlier integration \citep{wu2024semantic}. Furthermore, concurrent work show that visual representations arrive at layer zero already semantically dense, resembling the highly contextualized text found in later layers rather than raw, early-stage text tokens \citep{krojer2026latentlens}. Because visual inputs are pre-processed by a pre-trained vision encoder, they bypass the progressive, concept-building phase required for text. This informational density mismatch drives early layers to heavily focus on building textual context, deferring true cross-modal alignment to late-stage visual retrieval.

\section{Methodology}
To evaluate cross-modal alignment, we utilize a Visual Question Answering (VQA) retrieval task. We prompt the model with \texttt{"<image> How many objects are in the image?"} using 100-200 samples from the PixelProse dataset \citep{singla2024pixels} unless otherwise specified. Images were normalized, converted to RGB, and processed through a multimodal auto-processor. 

\textbf{Datasets}: We utilize the PixelProse dataset containing over 16 million image-caption pairs. For each entry, we use the image and its paired ground-truth caption. Inputs were formatted as a standard multimodal prompt sequence: \texttt{<image>} token followed by the human-made caption.

\textbf{Model}: We use each model's vision encoder and linear adapter that projects image patches into the language decoder's text space. The decoder consists of 18 layers (\textsc{PaliGemma}), each containing transformer blocks that integrate multi-modal information. Results for additional models can be found in the appendix. 

\subsection{Layer-wise Similarity Analysis \& Residual Stream Tracking}

To understand how visual embeddings evolve throughout the model, we track the internal representations of \textsc{PaliGemma-3B} across its 18 transformer layers. We sample the first 100 image-caption pairs. The image tokens are prepended to the text caption and delimited by a \texttt{<image>} token.  We apply a tensor mask to isolate the visual embeddings from their textual context. The \textsc{PaliGemma-3B} processor converts a $448 \times 448$ input image into exactly $N=1024$ discrete image patch tokens and applies an index mask to extract only these 1024 tokens at each layer $l$. The layer-wise similarity metric $S(l, l+1)$ over the \textsc{PixelProse} dataset is defined in Appendix~\ref{sec:cosine-metrics}.

To investigate the internal processing mechanisms of \textsc{PaliGemma-3B}, we also analyze the total magnitude of updates applied to the residual stream across all transformer layers. We standardized a dataset of multimodal prompts that alternated between text and image tokens. During the forward pass of these sequences, we register forward hooks at the final vector representation of each transformer block, so we can save the hidden states of the residual stream for every token across all layers. We then calculate the Euclidean Distance between a token's representation before entering a transformer block and immediately after exiting. To evaluate how the model routes different types of information, we isolate the text tokens from the vision tokens in the input and calculate the average update magnitude for each modality at each layer.


We further decompose the residual stream updates of vision tokens by passing them through the attention and MLP sublayers separately, isolating the relative contribution of each sublayer to visual processing across layers (Figure~\ref{fig:attention-vs-mlp-contributions-to-vision-tokens}).


\subsection{Activation Patching}
To determine how vision and text representations change through the network, we conduct an activation patching experiment evaluated via KL divergence. Interventions are performed within the shared residual stream of the language decoder, where each modality is isolated via strict token-index masking: the vision pathway targets the projected image tokens (indices 1-1024), while the textual pathway targets the prompt and generation tokens (indices 1025+). For each modality, we perform an initial baseline forward pass, caching the hidden activations from the early source layers (layers 2-4) and injecting them directly into the late layers (layers 15-17). The target layers' natural hidden states were overwritten with the cached early states. We quantified the model's shifts by calculating the KL divergence between the entire output vocabulary distribution of the clean run and each patched run.


\subsection{Attention Tracking \& Ablation}
Building on the activation patching results, we analyze how attention is distributed across modalities and layers to determine whether \textsc{PaliGemma-3B} integrates visual and textual information throughout the network or through late layers. For each layer, we extract the attention weights from the final token query and compute the mean attention towards vision and text tokens separately. 


To investigate whether attention from vision-dedicated heads causally influences model output, we perform an ablation study where we progressively suppress the top-$k$ vision heads, ranked by their mean attention to visual tokens. We compare three ablations --- $k \in \{5, 20, 50\}$ --- and evaluate the resulting degradation on two tasks: image captioning and visual question answering. To characterize how attention is redistributed under ablation, we visualize attention distributions as heatmaps across layers. Output quality is assessed using \textsc{PaliGemma-3B} prompted on both tasks, with responses evaluated by a VLM judge 
(\textsc{Gemma-3-12b-it}).

To determine which components drive these changes, we further decompose each layer's residual update by separately measuring the contributions of the attention and MLP sublayers, enabling a direct comparison of their roles in vision-token processing as shown in Figure \ref{fig:attention-vs-mlp-contributions-to-vision-tokens}. To determine which of these drives vision processing, we decompose each layer's residual stream contributions by passing vision tokens through the attention and MLP sublayers separately.



\subsection{kNN probing on Text and Vision}
We conduct a nearest-neighbor probe on the uncontextualized layer 0 embedding matrix for text tokens. For a set of diverse concepts (e.g., car, philosophy, ocean, running), we extract the baseline embedding vector and compute cosine similarities against the model's entire vocabulary. We identify the top-20 nearest neighbors for each concept and cluster them based on their similarity scores, defining a "semantic boundary" that marks the threshold at which lexical mapping degrades into noise.


To evaluate the semantic structure of visual representations before and after the linear adapter, we manually select an anchor image patch token and compute its top-5 nearest neighbors by cosine similarity. We perform this comparison at two stages: at the start of the vision encoder (pre-projection) and at the output of the linear adapter (post-projection). By examining how the anchor's neighborhood changes across these two stages, we assess whether projection into the language decoder's embedding space preserves the semantic structure of visual representations.

\subsection{Vision Token to Text Token Replacement }

To test whether high semantic-density text representations influence cross-modal attention, we identify 8 semantically meaningful caption tokens for each sample. For each selected text token, we find the most similar vision token using cosine similarity in PaliGemma's layer-0 latent space. We then replace these vision token vectors at layer-0 with their corresponding text tokens, leaving all other vision tokens unchanged. Attention is then measured before and after replacement at these token positions.

Results are visualized as scatter plots comparing attention values before and after replacing vision token positions with their nearest text equivalents, across early, intermediate, and late layers. Each point represents a token, positioned relative to the 
$y = x$ diagonal: tokens along the diagonal exhibit no change in received attention following replacement, tokens above it receive \textit{more} attention when the position is occupied by a text token, and tokens below it receive \textit{less}. Significant deviation from the diagonal thus indicates that the modality occupying a given position influences how other tokens attend to it.

\section{Experimental Results}

\subsection{Vision representations don't change across layers}

To quantify representational shifts, we computed the mean cosine similarity between isolated visual tokens. As shown in Figure~\ref{fig:consecutive-cosine-similarity}, the trajectory of visual representations across layers reveals two distinct phases:

\begin{figure}[htbp]
    \centering
    \includegraphics[width=\linewidth]{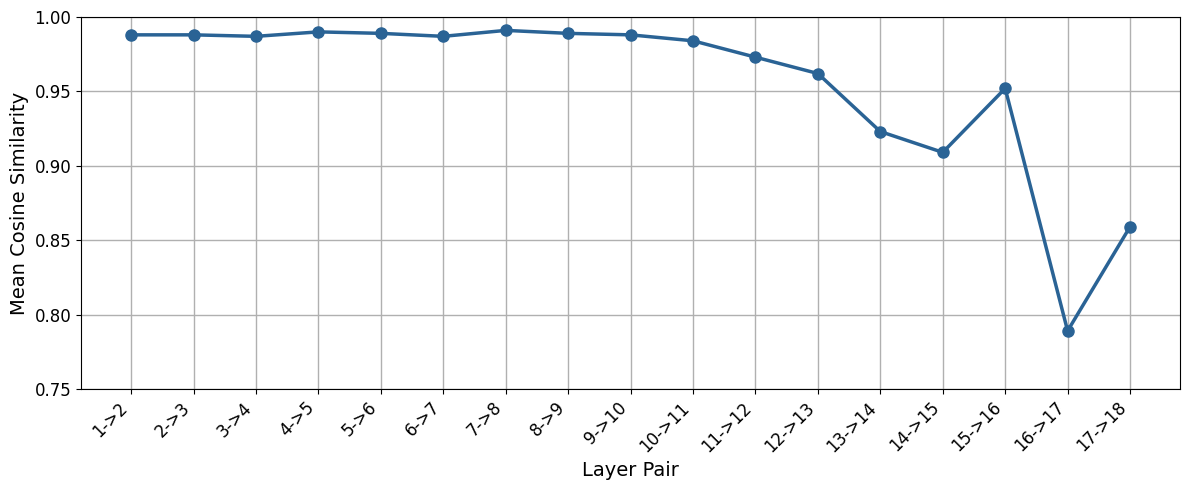}
    \Description{Evolution of Mean Cosine Similarity for Vision-Token Representations Across Successive Layers: The line plot tracks mean cosine similarity across transformer layer pairs. The similarity remains flat near $0.99$ through pair $9\rightarrow10$, steadily declines to approximately $0.91$ by pair $14\rightarrow15$, and drops sharply to a minimum of $0.79$ at pair $16\rightarrow17$ before a slight final increase.}
    \caption{Evolution of Mean Cosine Similarity for Vision-Token Representations Across Successive Layers: The line plot tracks mean cosine similarity across transformer layer pairs. The similarity remains flat throughout early and mid layers then slightly declines in later layers as cross-modal integration occurs.}
    \label{fig:consecutive-cosine-similarity}
\end{figure}

\textbf{Stable Layers (1--10):} Cosine similarity remains consistently high, indicating that the model preserves the core visual structure throughout early processing. Representations evolve gradually, with the network refining features and accumulating context while leaving the underlying visual geometry largely intact. 

\textbf{Transformation Layers (11--18):} Starting from layer 11, similarity begins a gradual decline, falling to $\approx 0.91$ by pair $14 \rightarrow 15$. This progressive shift reflects the model transitioning towards cross-modal integration: visual representations are increasingly restructured to align with the textual space, ultimately enabling token prediction from both modalities.

\begin{figure}[htbp]
\centering
\begin{subfigure}{.48\linewidth}
  \centering
  \includegraphics[width=\linewidth]{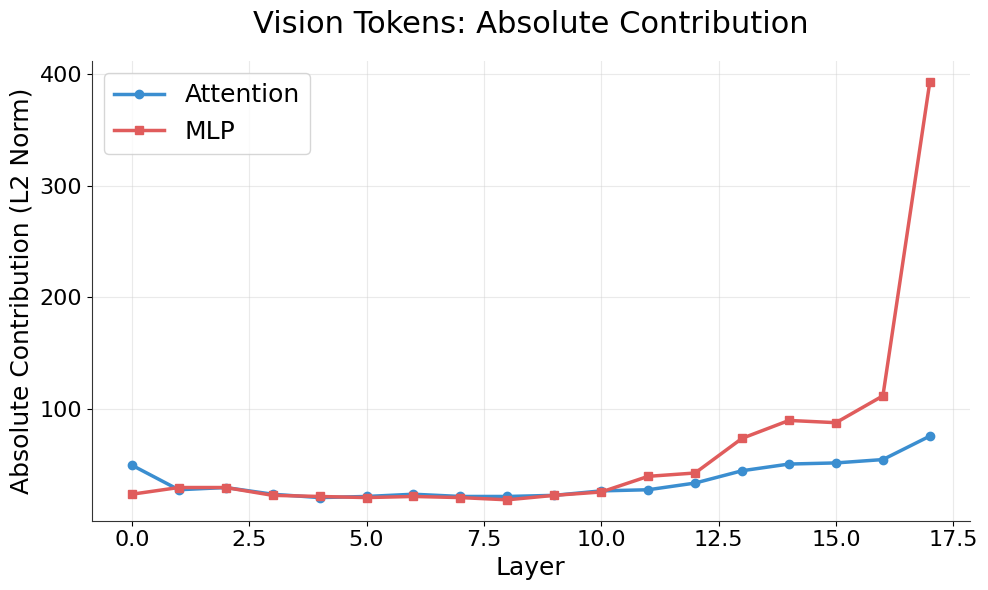}
  \caption{Vision Tokens Absolute Contributions}
  \label{fig:sub1}
\end{subfigure}\hfill
\begin{subfigure}{.48\linewidth}
  \centering
  \includegraphics[width=\linewidth]{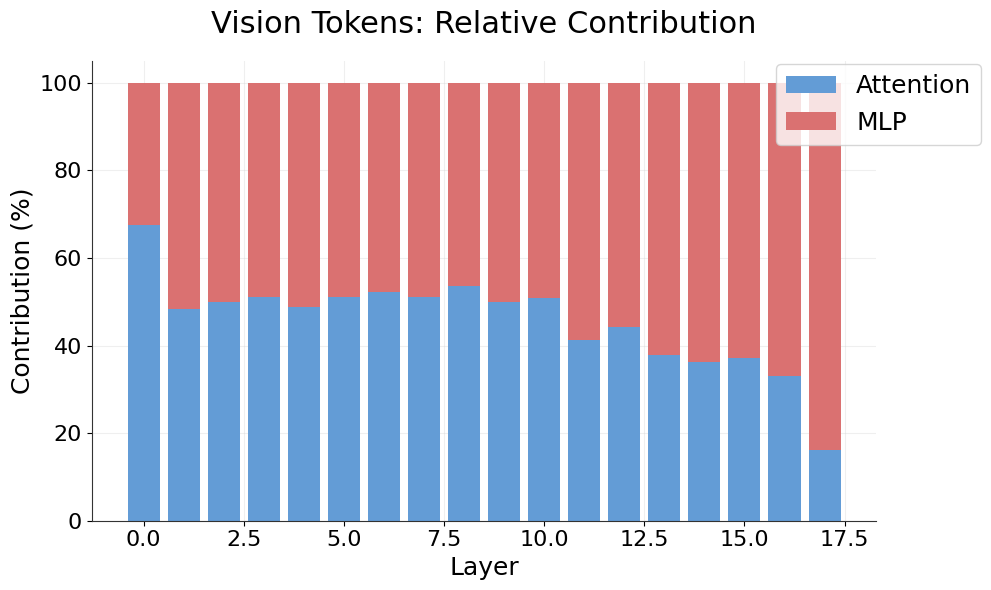}
  \caption{Vision Tokens Relative Contributions}
  \label{fig:sub2}
\end{subfigure}
\caption{Attention and MLP contributions to vision token updates: Relative contributions reveal attention and MLP contribute roughly equally throughout most layers, with attention comprising 40-60\% of total changes. MLP spikes at the final layer, comprising over 80\% of contributions to the residual stream. Absolute contributions demonstrate that the L2 Norm for MLP jumps from around 100 to 400 in the final transformer layer.}
\label{fig:attention-vs-mlp-contributions-to-vision-tokens}
\end{figure}

Furthermore, when early visual representations were patched directly into the late layers, the resulting distribution nearly mirrored the natural baseline, proving that vision representations undergo little to no transformation across layers. However, patching these same early visual tokens into the mid-layers produced a slight, moderate divergence ($\mu = 0.8881, \sigma = 0.5935$). Rather than indicating a change in the visual concept, the fact that the vision stream regains near-perfect stability in the final layers confirms that the visual representations themselves are unaltered.

Overall, our results show that textual representations in \textsc{PaliGemma-3B} are continuously refined across layers, while visual representations remain relatively stable and are only integrated later in computation. Visual information does not appear to act as an independent stream. Instead, it mainly serves as a signal that influences the reasoning process of the textual stream. As a result, later visual-layer activations cannot independently determine the model's final prediction and must first align with the language-processing representations before they can meaningfully affect the output.

\begin{figure}[htbp]
   \centering
   \includegraphics[width=\linewidth]{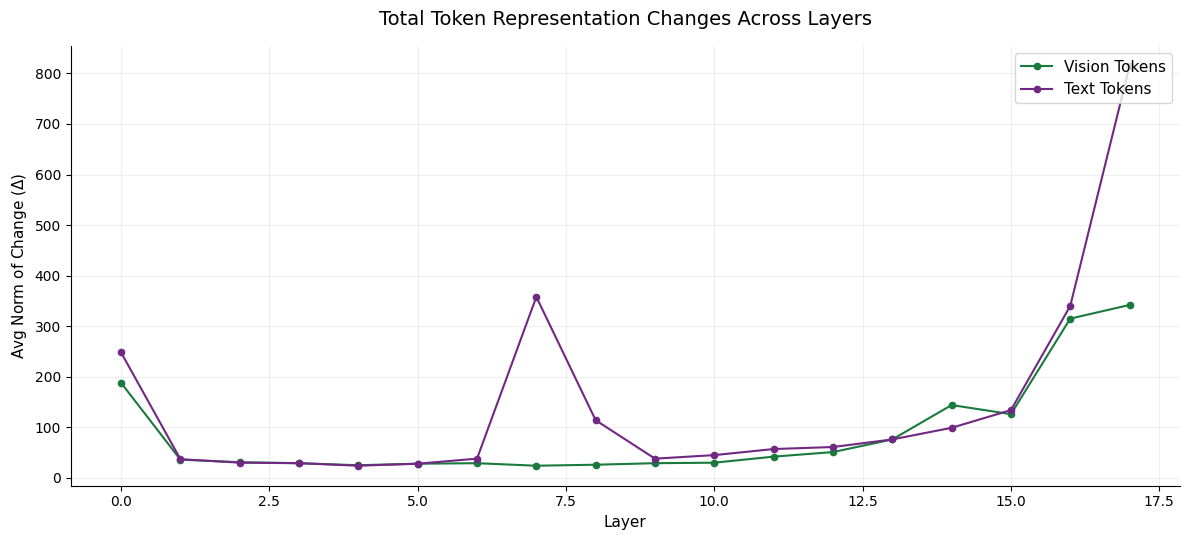}
    \caption{\textbf{Total Token Representation Changes Across Layers:} The line graph plots the average $L_{2}$ norm of change for vision and text tokens across transformer layers. Both lines show an initial spike at Layer 0, then remain low through Layer 6. The text token line spikes sharply at Layer 7 while the vision line remains flat. Both lines gradually increase from layers 8 to 14, followed by a steep rise in layers 15 to 17.}   
    \label{fig:vision-token-changes-per-layer}
\end{figure}

Finally, the residual stream updates (Figure \ref{fig:vision-token-changes-per-layer}) demonstrate the architectural bottleneck. \textbf{Layers 0-6: } Both modalities exhibit a sharp initial update at Layer 0, immediately followed by a step drop to a stable pattern through Layer 6. The initial spike is a standard architectural behavior that reflects the projection of raw token embeddings into the model's latent space. \textbf{Layer 7: } A significant structural change happens at Layer 7, where text tokens undergo a drastic representation shift, while visual tokens remain untouched. This modality asymmetry acts as a text-processing bottleneck, in which the model queries textual representations over the now-stabilized visual context. \textbf{Layers 8-14: } Moving through the middle layers, the update trajectories for both modalities begin to align and gradually drift upward, implying coupled processing. \textbf{Layers 15-17: } In the final layers, the update magnitudes escalate exponentially. However, at Layer 17, text tokens undergo a much more significant update than vision tokens. This reflects the model's behavior to align the residual stream with the final unembedding matrix. 

\begin{figure}[htbp]
    \centering
    \includegraphics[width=\linewidth, height=6cm]{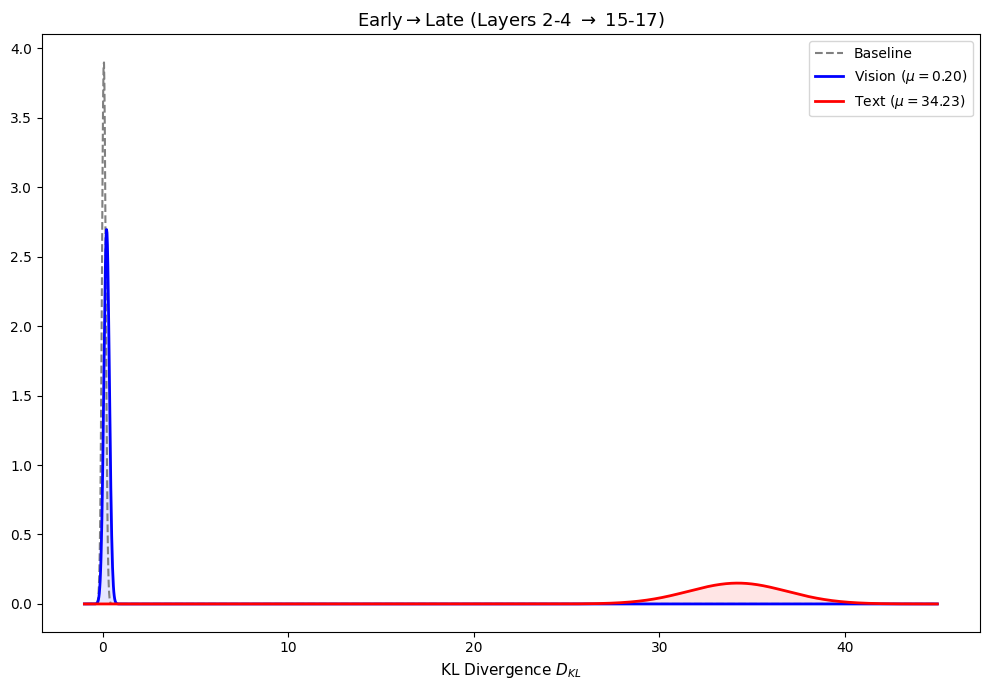}
    \caption{The density plot compares the KL Divergence of output distributions when early layer activations (2--4) are patched into late layers (15--17) in \textsc{PaliGemma-3B}. The resulting text intervention distribution is centered at $\mu=34.23$, while the vision intervention distribution is centered near zero at $\mu=0.20$.}    \label{fig:vision-text-kl}
\end{figure} 


\begin{figure}[htbp] 
    \centering
    \includegraphics[width=\linewidth]{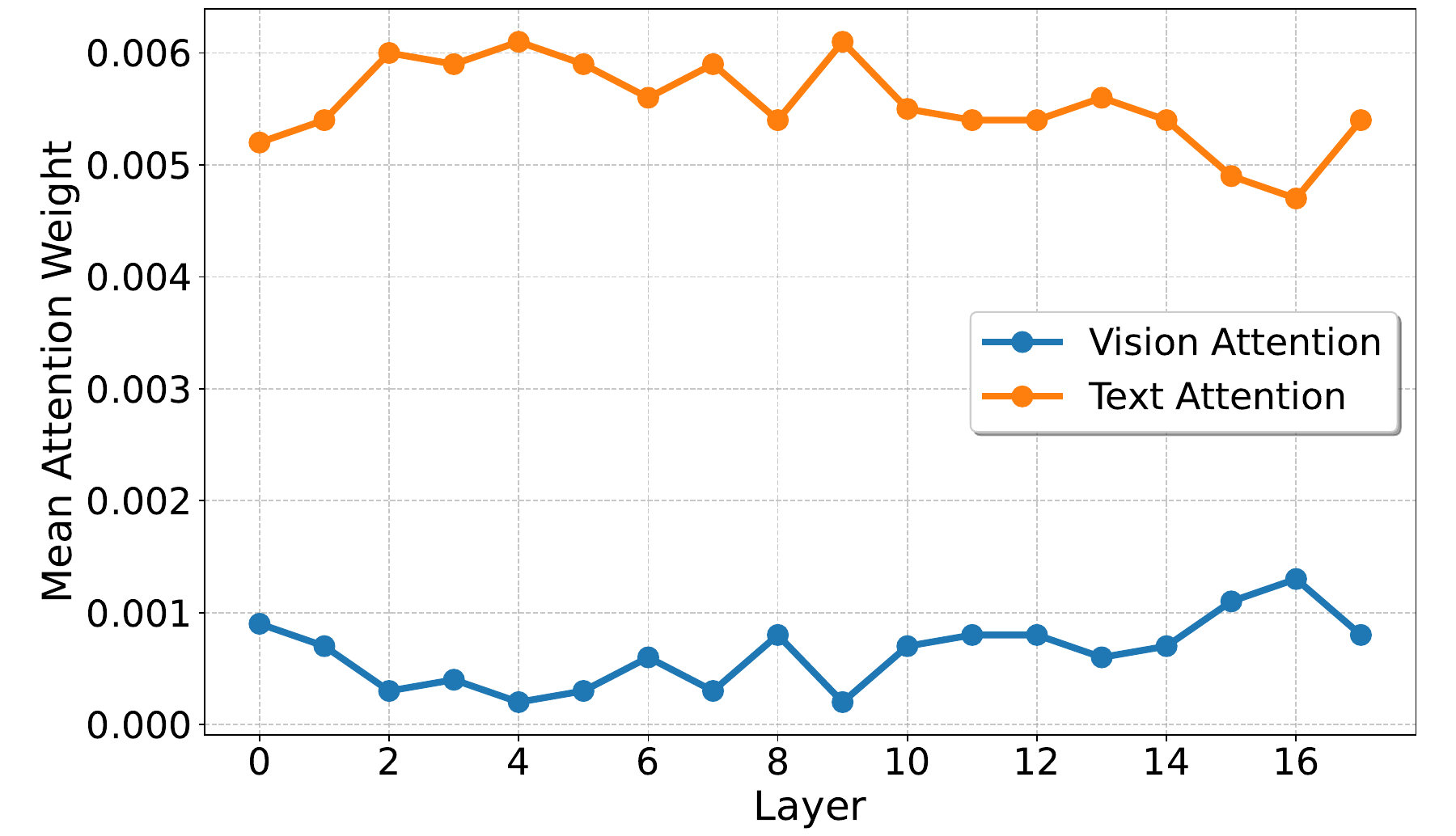}
    \caption{\textbf{Mean Vision vs. Text Attention Weights Across Layers:} The plot displays mean attention weights for vision and text tokens across transformer layers 0--17. The text attention line fluctuates between 0.0047 and 0.0061, while the vision attention line ranges from 0.0002 to 0.0009 before rising to a peak of 0.0013 at layer 16.}    \label{fig:modality-attention-weights}
\end{figure}
As shown in Figure \ref{fig:modality-attention-weights}, the model shows a clear reliance on textual tokens in most layers, with visual attention increasing only in the final layers. This indicates that \textsc{PaliGemma} defers integrating visual features for the final layers, where cross-modal alignment for answer generation occurs. 

\begin{figure}[ht!]
   \centering
   \includegraphics[width=\linewidth]{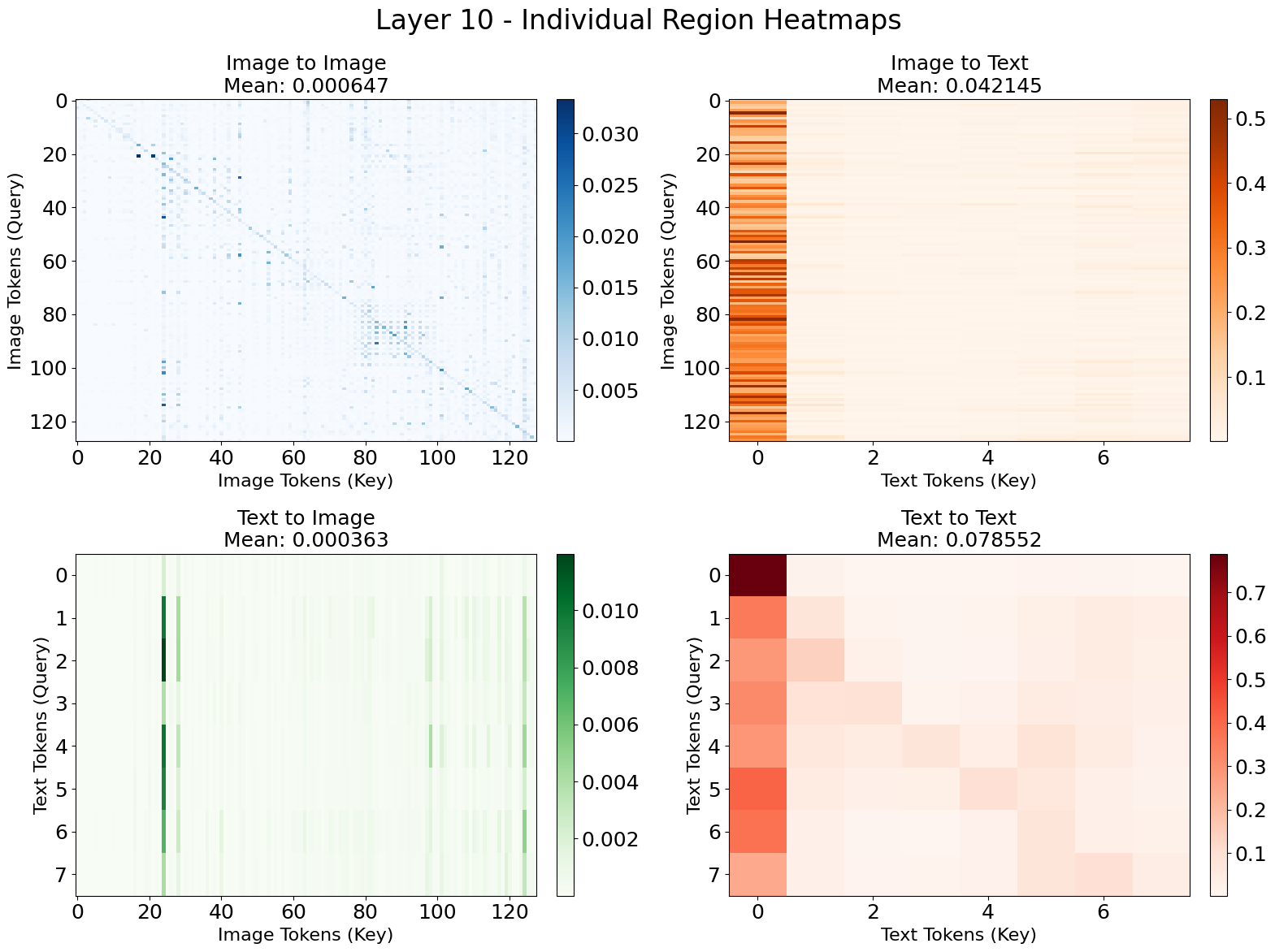}
   \caption{Attention score decomposition for \textsc{PaliGemma-3B}: Each quadrant shows attention from query tokens (rows) to key tokens (columns). Top-left: Vision queries attending to vision keys (V to V, mean=0.000647). Top-right: Vision queries attending to text keys (V to T, mean=0.000363). Bottom-left: Text queries attending to vision keys (T to V, mean=0.042145). Bottom-right: Text queries attending to text keys (T to T, mean=0.078552).}
   \label{fig:attention-scores-sub-contributions}
\end{figure}

Figure \ref{fig:attention-scores-sub-contributions} shows that image token queries primarily attend to image token keys rather than incorporating information from text queries. Similarly, text token queries primarily attend to text token keys, with low attention from image queries. This aligns with the "process-then-retrieve" workflow, which defers information integration across modalities in intermediate layers. Cross-modal attention is negligible in early layers and becomes significant only in the final transformer blocks.

In our ablation trials, the 5-head ablation trial showed captioning quality degrading. In the 10-head ablation trial, captioning quality degrades measurably, confirming that attention becomes less focused on task-relevant image regions. The heatmaps illustrate this: in the baseline, high-attention tokens cluster around the primary subject, while the ablated run distributes weight diffusely across the image. In the 50-head ablation trial, degradation is more severe, and the model begins producing captions that rely heavily on generic language priors. 

Importantly, the degradation pattern under questioning differs from the captioning condition in one respect: the model more frequently produces answers that are linguistically plausible but visually incorrect (e.g., naming a common object category rather than the specific subject present). This aligns with our patching results, which show that textual representations dominate the early and middle layers; once vision heads are suppressed, the model's answer generation falls back on these linguistic priors with minimal grounding from the image.

Across captioning trials, the VLM judge rated baseline accuracy as \textit{Yes} in 83\% of cases, with 81\% for 5-head ablation, 78\% 10-head ablation and 66\% for 20-head ablation. For the questioning task, baseline accuracy as rated by the judge stood at 68\%, with a sudden increase to 72\% under both 5- and 10-head ablation. At 20-head ablation, the answer accuracy drops once more to 68\%. These results provide causal evidence that the removal of top-contributing vision attention heads measurably degrades both the distributional focus of attention and the accuracy of downstream outputs.


\subsection{Vision tokens are information dense even at layer 0 as opposed to text tokens which acquire semantic information gradually}

\begin{figure}[htbp]
\centering

\begin{minipage}{\linewidth}
  \centering
  \includegraphics[width=\linewidth]{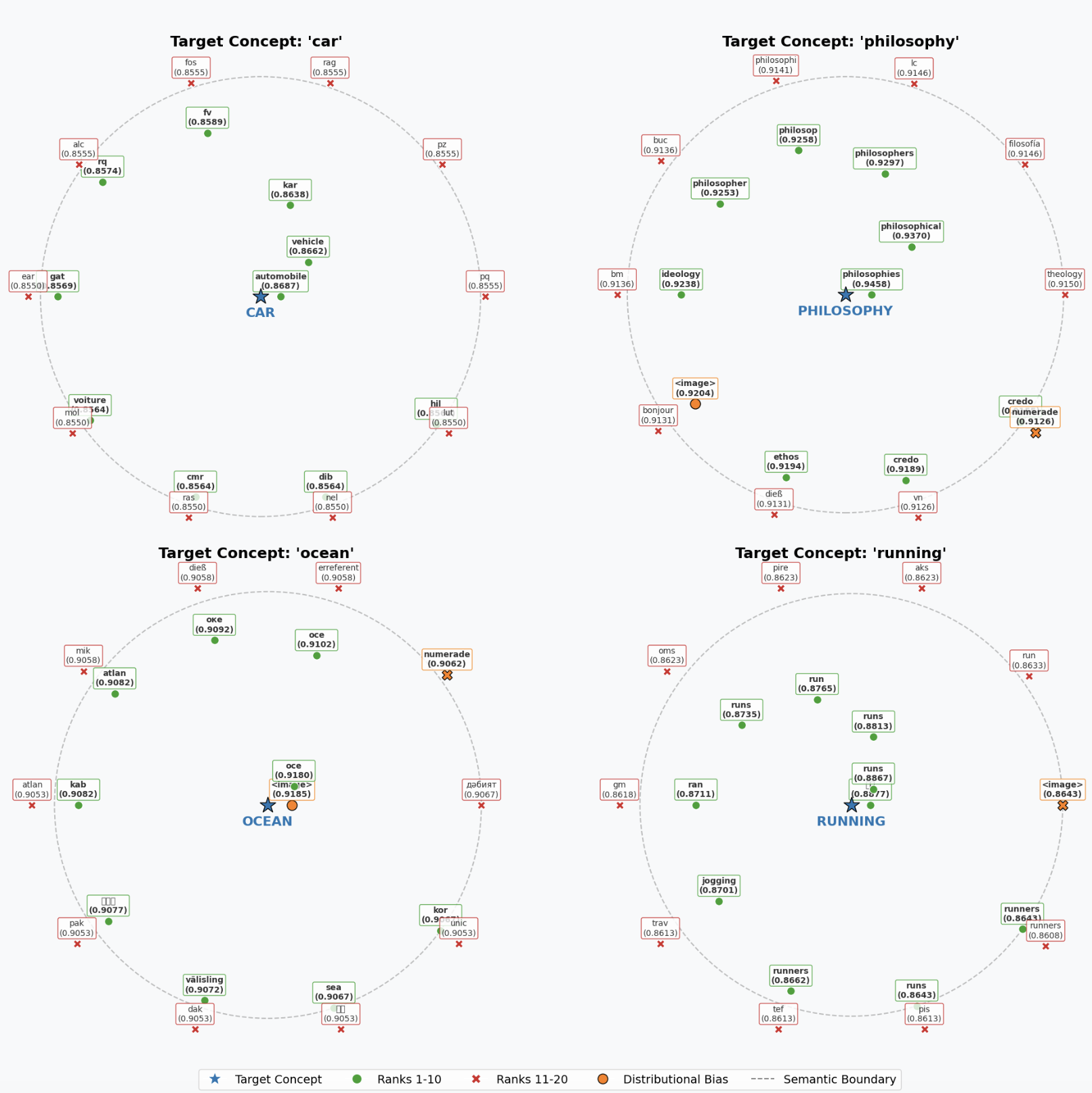}
\end{minipage}

\caption{\textbf{Semantically Relevant Tokens and Corresponding Neighbors:} The plots map uncontextualized embedding vectors for four target concepts (`car', `philosophy', `ocean', and `running') to their top-20 nearest neighbors by cosine similarity, with a dashed line denoting a boundary and orange markers indicating specific tokens outside this threshold.}
\label{fig:text-knn}
\end{figure}

kNN probing of text tokens within the semantic text boundary reveals that the model fails to retrieve true semantic synonyms, instead organizing its neighbors around lexical variations. Immediately upon crossing this threshold, the vector space collapses into statistical noise, where target concepts are paired with meaningless sub-word tokens (e.g., pz, pq, rag) with identical cosine similarity scores. This degraded space exhibits severe distributional bias by linking conceptually unrelated tokens, such as scraped textbook websites (e.g., Numerade) and functional prompt tokens (e.g., <image>), with standard vocabulary as shown in Figure \ref{fig:text-knn}.

\begin{figure}[htbp]
\centering

\begin{minipage}{\linewidth}
  \centering
  \includegraphics[width=\linewidth]{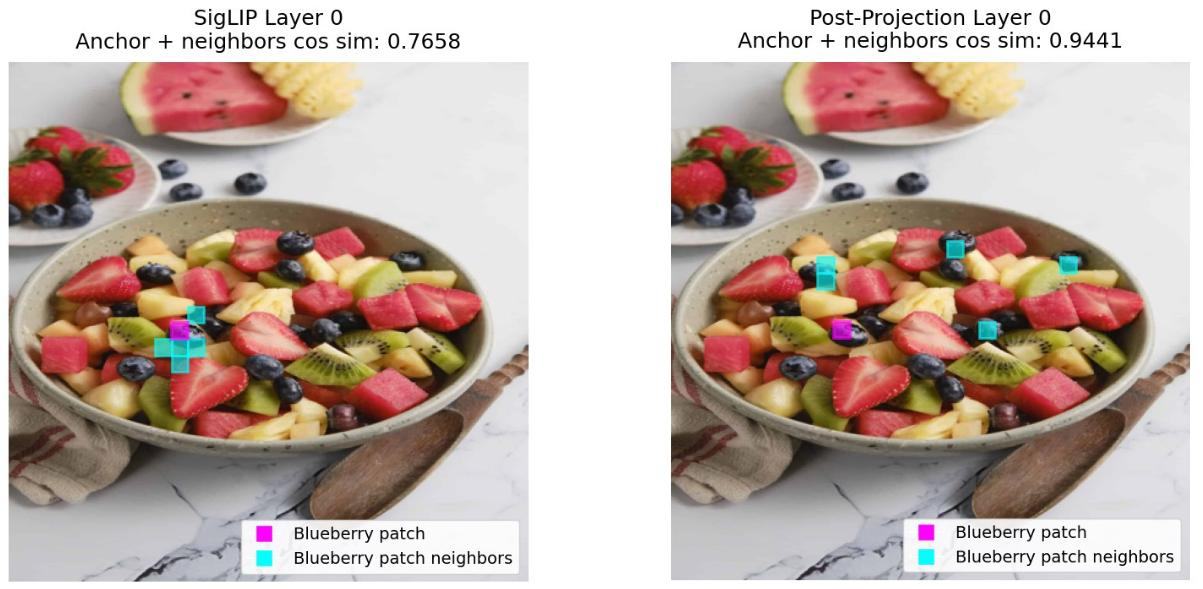}
  \centering
  \includegraphics[width=\linewidth]{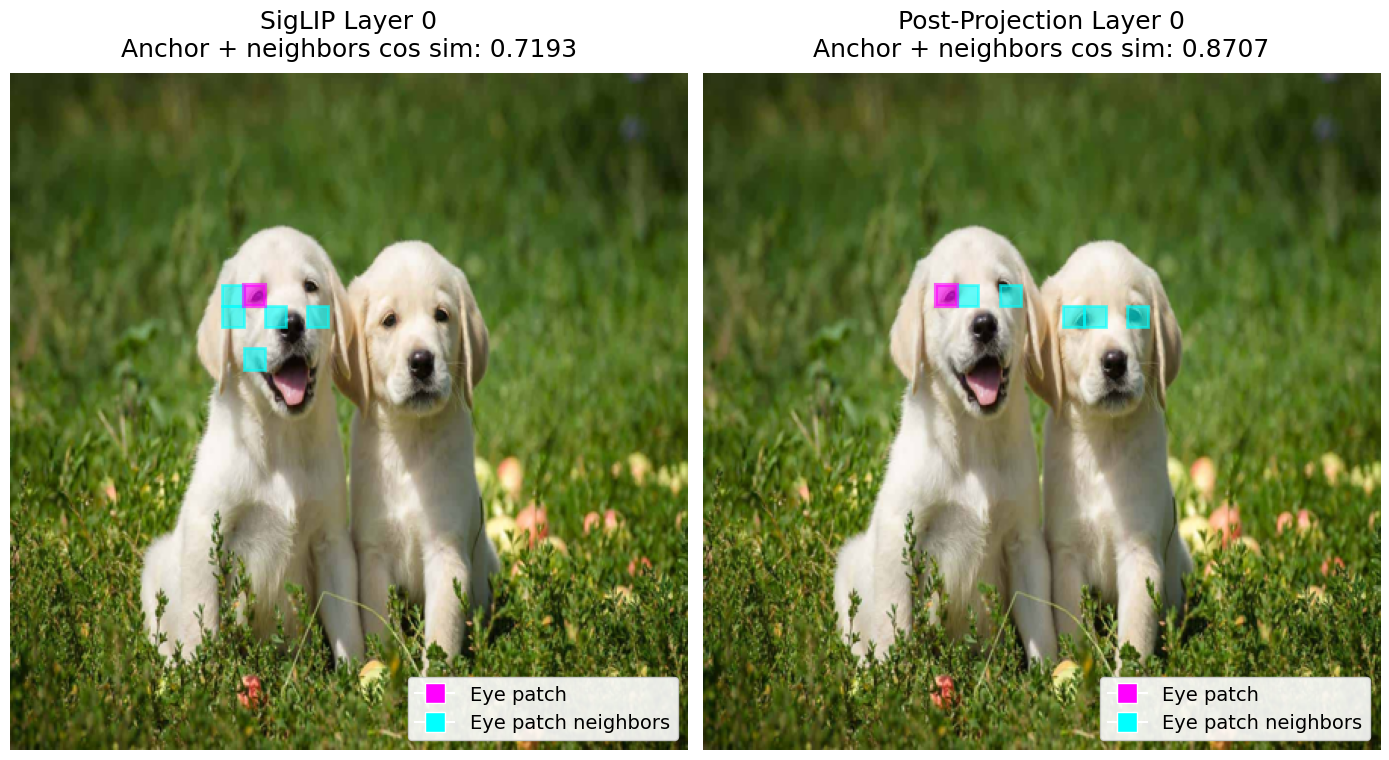}
\end{minipage}

\caption{\textbf{Semantically Relevant Tokens and Corresponding Neighbors:} The top photo of a fruit bowl is assigned an image patch of a blueberry. The bottom photo of two dogs is assigned an image patch of an eye. The 5-nearest neighbors are taken before and after projection at layer 0.}
\label{fig:vision-knn}
\end{figure}

In contrast, vision-token behavior follows a different pattern. Rather than acquiring meaning progressively, visual embeddings enter the model already semantically structured. As shown in Figure~\ref{fig:vision-knn}, nearest neighbors at layer 0 already exhibit coherent semantic similarity, with image patches aligning to visually related regions indicating that meaningful representational structure is present from the outset. This early semantic organization is directly supported by \citet{krojer2026latentlens}, who demonstrate that layer-zero visual representations more closely resemble the highly contextualized token representations found in the later layers of the model than those found in early layers.



Because visual inputs are pre-processed by a pre-trained vision encoder, they bypass the progressive, data processing phase required for text. Together, these results prove that, unlike text, vision tokens arrive semantically dense at layer zero, occupying highly organized subspaces of the latent space prior to any cross-modal integration.

\begin{figure}[htbp]
    \centering
    \includegraphics[width=\linewidth]{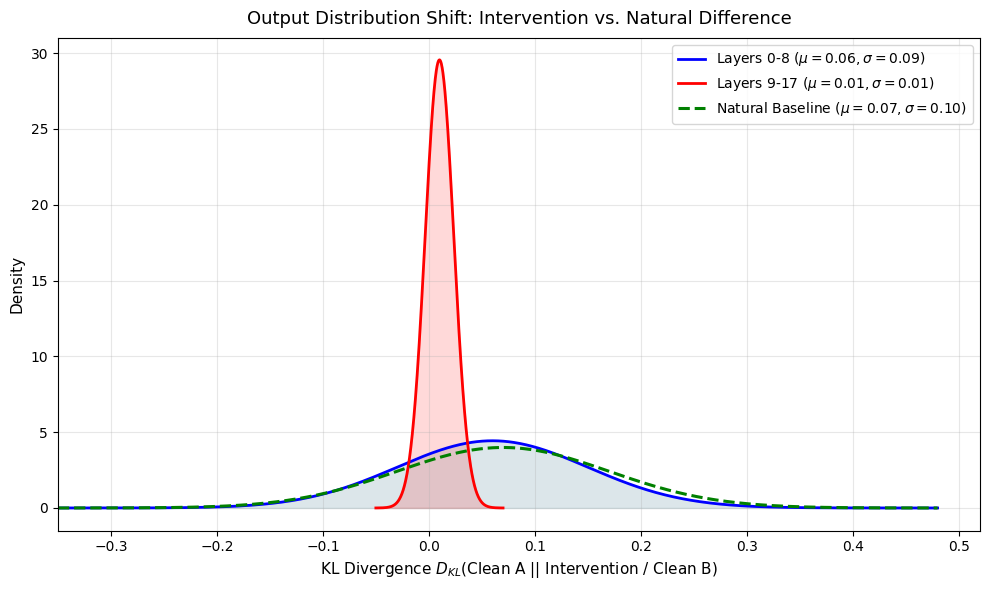}
    \caption{\textbf{Output Distribution Shifts Measured via KL Divergence for \textsc{PaliGemma-3B}:} The density plot displays the natural baseline output variation ($\mu=0.07, \sigma=0.10$) alongside interventions where visual activations are patched into early layers (0--8) ($\mu=0.06, \sigma=0.09$) and late layers (9--17) ($\mu=0.01, \sigma=0.01$).}    \label{fig:kl-divergence}
\end{figure}

\subsection{High token semantic density influences the formation of the modality gap}

\begin{figure}[htbp]
    \centering
    \includegraphics[width=\linewidth]{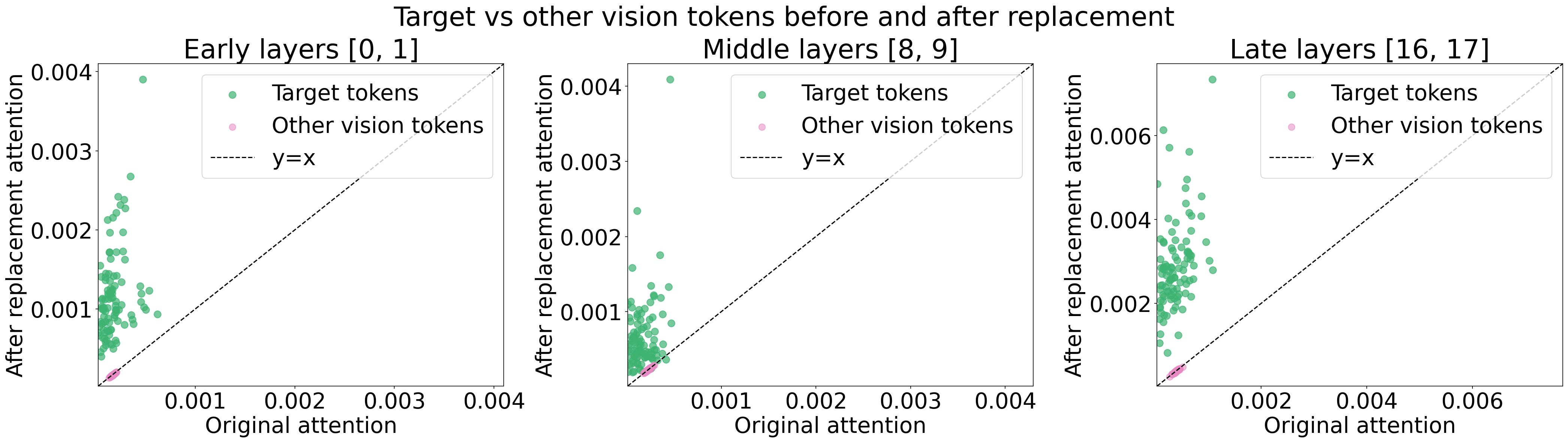}
    \caption{Traced 8 target tokens before and after text token replacement for \textsc{PaliGemma-3B}: Each dot represents a sample out of the 100 samples run. In all three graphs, attention is higher after the vision token target position is replaced with a text token, but the change varies. In early layers, attention to replaced positions is mostly condensed between 0.0005 to 0015. In intermediate layers, attention to replaced positions mostly stays under 0.001. In late layers, the attention to replaced tokens is much higher with values ranging between 0.002 to 0.004.}
    \label{fig:vision-token-replacement}
\end{figure} 

Replacing the target vision tokens with their semantically closest text tokens substantially increases attention to those positions across all layer groups, with peaks at early and late layers as shown in Figure \ref{fig:vision-token-replacement}. Since the replaced tokens preserve the original semantic content, the increase in attention cannot be attributed to a change in meaning. Instead, we interpret this as the model responding to the tokens' modality: text tokens, lacking the pre-built semantic structure of vision tokens, require more data processing across layers. By substituting text-space equivalents at layer zero, the model treats those positions as semantically immature, allocating greater attention accordingly. These results support the view that the reduced attention received by vision tokens is a direct consequence of their early semantic density. The model recognizes that little further processing is needed and allocates attention.


\section{Discussion}

In this work, we present a mechanistic analysis of adapter-based vision-language models (VLMs) to investigate the origins of the modality gap and understand how visual information is integrated throughout the network. Our central hypothesis is that mismatches in semantic density between visual and textual representations at the embedding layer contribute to the persistent underutilization of visual information in multimodal reasoning. While prior work has observed that visual tokens receive consistently lower attention compared to text tokens, the underlying mechanisms behind this imbalance remain poorly understood. Our analysis provides mechanistic evidence showing that many VLMs implicitly adopt a "process-then-retrieve" workflow, where textual representations dominate the computation while visual representations remain largely dormant until the final layers.

Our findings raise fundamental questions about the efficiency of current multimodal representation learning paradigms. Across early and middle layers, we observe strong textual processing dynamics alongside comparatively static visual representations, indicating that the two modalities are not integrated symmetrically. Instead, the model appears to continuously refine contextual representations. This delayed integration offers a mechanistic explanation for why many VLMs struggle on vision-centric reasoning tasks. Rather than jointly reasoning over both modalities throughout the forward pass, the model prioritizes linguistic priors first and only incorporates visual evidence at later stages, even when that evidence contradicts the textual context. Through kNN probing and activation patching, we identify where these multimodal misalignment emerge and how they propagate across layers.

Our cosine similarity analysis further supports this interpretation. In \textsc{PaliGemma-3B}, vision-token representations remain highly stable across layers before exhibiting a sharp representational transition between layers 11 and 18. This transition likely corresponds to the point at which visual features are finally projected into the textual latent space and become usable by downstream language-processing circuits. The stability of visual representations in earlier layers suggests that image tokens are treated more as static conditioning vectors than actively processed semantic representations.

Additionally, the early-to-late patching experiments provide further evidence for this asymmetric computation pattern. As shown in Appendix Section~\ref{sec:paligemma2b-text-front}, early-to-late textual patching fails because the textual stream depends on iterative refinement across the full network depth to construct the final prediction. In contrast, the success of early-to-late vision patching, reflected by positive $\Delta$ logit values, indicates that visual information is preserved throughout the network despite remaining largely inactive. This suggests that the limitation is not the disappearance of visual features, but rather the delayed retrieval and utilization of those features by the language model.

These findings point toward several promising architectural directions for reducing the modality gap. Future work should investigate mechanisms that encourage earlier cross-modal interaction. More broadly, our results suggest that improving multimodal reasoning may require moving away from architectures that treat vision as a late-stage signal and toward systems that perform continuous joint reasoning across modalities throughout the forward computation. Understanding the tradeoff between earlier visual integration and language modeling remains an important open problem for the design of future VLM architectures.

\section{Conclusion}
In this work, we analyzed the mechanistic structure of adapter-based VLMs, using \textsc{PaliGemma-3B}, \textsc{PaliGemma2-10B}, and \textsc{Qwen2-VL} to confirm that these architectures exhibit early-layer textual data processing followed by late-stage visual information retrieval and high vision token semantic density at layer zero. Our findings raise fundamental questions about current vision-language architectures. The persistent separation of modalities until the final transformer blocks indicates that current architectures rely on modality-specific processing streams that only merge near the output layer. The dominance of textual attention across the network, with cross-modal visual attention peaking only during final answer generation, reveals an unbalanced architectural focus on textual biases. Furthermore, our patching experiments confirm that deferring visual integration creates a computational bottleneck, restricting the network's capacity for complex visual reasoning. Ultimately, our mechanistic analysis suggests that developing VLMs with true multi-modal understanding will require architectural shifts that enforce deeper, earlier cross-modal alignment.

\section{Limitations}



While our analysis addresses a key gap in understanding cross-modal integration, several limitations remain.

\textbf{Statistical Rigor:} Our analyses are conducted over 100--200 samples. Additionally, we do not systematically vary image complexity or question type, leaving open the question of how representational patterns may shift across different input tasks.

\textbf{Architectural Scope:} All experiments are conducted on adapter-based VLMs (\textsc{PaliGemma-3B}~\cite{beyer2024paligemma}, \textsc{PaliGemma2-10B}~\cite{keysers2024introducing}, and \textsc{Qwen2-VL-2B}~\cite{bai2023qwen}). Although \textsc{PaliGemma2-10B} offers some insight into scaling behavior, all evaluated models remain relatively modest in scale. It is unclear whether the observed patterns generalise to state-of-the-art large-scale VLMs or to architectures that differ fundamentally in design, such as contrastive or unified multimodal models.

\textbf{Causal Validation:} While we identify the process-then-retrieve pattern and attribute it to the structural constraints imposed by adapter-based models, our analysis does not verify this hypothesis through controlled retraining experiments that ablate or modify the integration mechanism directly. Future work should aim to move beyond observation toward intervention.

\section{Author Contributions}
Arpita Shanbag, Julia Train, and Dhruv Mandala contributed to the implementation of the research, analysis of the results, and wrote the manuscript under the consultation of Ayda Sultan. 

Arpita Shanbhag conducted layer-wise similarity analysis and residual stream tracking (Section 3.1), as well as the activation patching experiments (Section 3.2). Furthermore, A.S. led the text-specific kNN probing on the embedding matrix (Section 3.4) and was responsible for generating Figures 1, 3, 4, 5, 7, and 9.

Julia Tran conducted attention tracking and ablation (Section 3.3), image-specific kNN probing, and vision token to text token replacement (Section 3.5). J.T. was responsible for generating Figures 2, 5, 6, 8, and 10.

Ayda Sultan conceptualized the research question and overall direction of the project, designed the experimental methodology, and analyzed and interpreted the results. A.S also contributed to reviewing and editing the manuscript, helping shape the paper's framing and presentation.

\section{Acknowledgments}

We would like to thank Algoverse for supporting and funding this project. We are also grateful to Javier Ferrando for his valuable feedback on the experimental design and for reviewing multiple drafts of the paper.

\clearpage
\nocite{*}
\bibliographystyle{ACM-Reference-Format}
\bibliography{references}

@String(CVPR= {IEEE Conf. Comput. Vis. Pattern Recog.})

@String(NIPS= {Adv. Neural Inform. Process. Syst.})

@String(ICLR = {Int. Conf. Learn. Represent.})

@misc{keysers2024introducing,
 author = {Keysers, Daniel and Steiner, Andreas},
 title = {Introducing PaliGemma 2: Powerful Vision-Language Models, Simple Fine-Tuning},
 year = 2024,
 url = {https://developers.googleblog.com/en/introducing-paligemma-2-powerful-vision-language-models-simple-fine-tuning/},
 urldate = {2026-02-05}
}

@misc{shantanu2025tik,
 author = {Shantanu, Jain},
 title = {tiktoken},
 year = 2025,
 url = {https://github.com/openai/tiktoken/},
 urldate = {2026-02-01}
}

@article{jiang2023pre,
 title={Pre-RMSNorm and Pre-CRSNorm Transformers: Equivalent and Efficient Pre-LN Transformers},
 author={Jiang, Zixuan and Gu, Jiaqi and Zhu, Hanqing and Pan, David},
 journal=NIPS,
 volume={36},
 pages={45777--45793},
 year={2023}
}

@article{shazeer2020glu,
 title={GLU Variants Improve Transformer},
 author={Shazeer, Noam},
 journal={arXiv preprint arXiv:2002.05202},
 year={2020}
}

@article{bai2023qwen,
 title={Qwen technical report},
 author={Bai, Jinze and Bai, Shuai and Chu, Yunfei and Cui, Zeyu and Dang, Kai and Deng, Xiaodong and Fan, Yang and Ge, Wenbin and Han, Yu and Huang, Fei and others},
 journal={arXiv preprint arXiv:2309.16609},
 year={2023}
}

@article{shukor2024,
 title={Implicit Multimodal Alignment: On the Generalization of Frozen {LLMs} to Multimodal Inputs},
 author={Shukor, Mustafa and Cord, Matthieu},
 journal={arXiv preprint arXiv:2405.16700},
 year={2024}
}

@article{rajaram2025,
 title={Line of Sight: On Linear Representations in {VLLMs}},
 author={Rajaram, Achyuta and Schwettmann, Sarah and Andreas, Jacob and Conmy, Arthur},
 journal={arXiv preprint arXiv:2506.04706},
 year={2025}
}

@article{neo2024,
 title={Towards Interpreting Visual Information Processing in Vision-Language Models},
 author={Neo, Clement and Ong, Luke and Torr, Philip and Geva, Mor and Krueger, David and Barez, Fazl},
 journal={arXiv preprint arXiv:2410.07149},
 year={2024}
}

@inproceedings{sim2025,
 title={Can {VLMs} Actually See and Read? A Survey on Modality Collapse in Vision-Language Models},
 author={Sim, Mong Yuan and Zhang, Wei Emma and Dai, Xiang and Fang, Biaoyan},
 booktitle={Findings of the Association for Computational Linguistics: ACL 2025},
 pages={24452--24470},
 year={2025}
}

@article{liang2022,
 title={Mind the gap: Understanding the modality gap in multi-modal contrastive representation learning},
 author={Liang, Weixin and Zhang, Yuhui and Kwon, Yongchan and Yeung, Serena and Zou, James},
 journal=NIPS,
 volume={35},
 pages={17612--17625},
 year={2022}
}

@article{jiang2023,
 title={Mistral 7B},
 author={Jiang, Albert Q and Sablayrolles, Alexandre and Mensch, Arthur and Bamford, Chris and Chaplot, Devendra Singh and Casas, Diego de las and Bressand, Florian and Lengyel, Gianna and Lample, Guillaume and Saulnier, Lucile and others},
 journal={arXiv preprint arXiv:2312.06968},
 year={2023}
}

@inproceedings{tong2024,
 title={Eyes Wide Shut? Exploring the Visual Shortcomings of Multimodal {LLMs}},
 author={Tong, Shengbang and Liu, Zhuang and Zhai, Yuexiang and Ma, Yi and LeCun, Yann and Xie, Saining},
 booktitle=CVPR,
 pages={9568--9578},
 year={2024}
}

@article{venhoff2025,
 title={How Visual Representations Map to Language Feature Space in Multimodal {LLMs}},
 author={Venhoff, Constantin and Khakzar, Ashkan and Joseph, Sonia and Torr, Philip and Nanda, Neel},
 journal={arXiv preprint arXiv:2506.11976},
 year={2025}
}

@inproceedings{kang2025,
 title={See What You Are Told: Visual Attention Sink in Large Multimodal Models},
 author={Kang, Seil and Kim, Jinyeong and Kim, Junhyeok and Hwang, Seong Jae},
 booktitle=ICLR,
 year={2025},
 url={https://openreview.net/forum?id=7uDI7w5RQA}
}

@article{liu2025,
 title={Seeing but Not Believing: Probing the Disconnect Between Visual Attention and Answer Correctness in {VLMs}},
 author={Liu, Zhining and Chen, Ziyi and Liu, Hui and Luo, Chen and Tang, Xianfeng and Wang, Suhang and Zeng, Joy and Dai, Zhenwei and Shi, Zhan and Wei, Tianxin and others},
 journal={arXiv preprint arXiv:2510.17771},
 year={2025}
}

@article{venhoff2025too,
 title={Too Late to Recall: Explaining the Two-Hop Problem in Multimodal Knowledge Retrieval},
 author={Venhoff, Constantin and Khakzar, Ashkan and Joseph, Sonia and Torr, Philip and Nanda, Neel},
 journal={arXiv preprint arXiv:2512.03276},
 year={2025},
 url={https://openreview.net/forum?id=qeL8fi8GS7}
}

@article{nikankin2025same,
 title={Same Task, Different Circuits: Disentangling Modality-Specific Mechanisms in VLMs},
 author={Nikankin, Yaniv and Arad, Dana and Gandelsman, Yossi and Belinkov, Yonatan},
 journal={arXiv preprint arXiv:2506.09047},
 year={2025}
}

@misc{merullo2023linearlymappingimagetext,
 title={Linearly Mapping from Image to Text Space}, 
 author={Jack Merullo and Louis Castricato and Carsten Eickhoff and Ellie Pavlick},
 year={2023},
 eprint={2209.15162},
 archivePrefix={arXiv},
 primaryClass={cs.CL},
 url={https://arxiv.org/abs/2209.15162}, 
}

@article{papadimitriou2025interpreting,
 title={Interpreting the linear structure of vision-language model embedding spaces},
 author={Papadimitriou, Isabel and Su, Huangyuan and Fel, Thomas and Kakade, Sham and Gil, Stephanie},
 journal={arXiv preprint arXiv:2504.11695},
 year={2025}
}

@inproceedings{jiang2024hallucination,
 title={Hallucination augmented contrastive learning for multimodal large language model},
 author={Jiang, Chaoya and Xu, Haiyang and Dong, Mengfan and Chen, Jiaxing and Ye, Wei and Yan, Ming and Ye, Qinghao and Zhang, Ji and Huang, Fei and Zhang, Shikun},
 booktitle=CVPR,
 pages={27036--27046},
 year={2024}
}

@article{wu2024semantic,
 title={The semantic hub hypothesis: Language models share semantic representations across languages and modalities},
 author={Wu, Zhaofeng and Yu, Xinyan Velocity and Yogatama, Dani and Lu, Jiasen and Kim, Yoon},
 journal={arXiv preprint arXiv:2411.04986},
 year={2024}
}

@article{schrodi2024two,
 title={Two Effects, One Trigger: On the Modality Gap, Object Bias, and Information Imbalance in Contrastive Vision-Language Models},
 author={Schrodi, Simon and Hoffmann, David T and Argus, Max and Fischer, Volker and Brox, Thomas},
 journal={arXiv preprint arXiv:2404.07983},
 year={2024}
}

@inproceedings{eslami2025mitigate,
 title={Mitigate the gap: Improving cross-modal alignment in CLIP},
 author={Eslami, Sedigheh and de Melo, Gerard},
 booktitle=ICLR,
 year={2025}
}

@article{luo2024vision,
 title={Vision-language models create cross-modal task representations},
 author={Luo, Grace and Darrell, Trevor and Bar, Amir},
 journal={arXiv preprint arXiv:2410.22330},
 year={2024}
}

@inproceedings{cohen2025performance,
 title={Performance gap in entity knowledge extraction across modalities in vision language models},
 author={Cohen, Ido and Gottesman, Daniela and Geva, Mor and Giryes, Raja},
 booktitle={Proceedings of the 63rd Annual Meeting of the Association for Computational Linguistics (Volume 1: Long Papers)},
 pages={29095--29108},
 year={2025}
}

@article{fu2025hidden,
 title={Hidden in plain sight: VLMs overlook their visual representations},
 author={Fu, Stephanie and Bonnen, Tyler and Guillory, Devin and Darrell, Trevor},
 journal={arXiv preprint arXiv:2506.08008},
 year={2025}
}

@article{hua2025vision,
 title={How Do Vision-Language Models Process Conflicting Information Across Modalities?},
 author={Hua, Tianze and Yun, Tian and Pavlick, Ellie},
 journal={arXiv preprint arXiv:2507.01790},
 year={2025}
}

@article{role2025fill,
 title={Fill the Gap: Quantifying and Reducing the Modality Gap in Image-Text Representation Learning},
 author={Role, Fran{\c{c}}ois and Meyer, S{\'e}bastien and Amblard, Victor},
 journal={arXiv preprint arXiv:2505.03703},
 year={2025}
}

@article{fahim2024s,
 title={It's Not a Modality Gap: Characterizing and Addressing the Contrastive Gap},
 author={Fahim, Abrar and Murphy, Alex and Fyshe, Alona},
 journal={arXiv preprint arXiv:2405.18570},
 year={2024}
}

@article{wang2024picture,
 title={Is a picture worth a thousand words? delving into spatial reasoning for vision language models},
 author={Wang, Jiayu and Ming, Yifei and Shi, Zhenmei and Vineet, Vibhav and Wang, Xin and Li, Sharon and Joshi, Neel},
 journal=NIPS,
 volume={37},
 pages={75392--75421},
 year={2024}
}

@article{singla2024pixels,
 title={From pixels to prose: A large dataset of dense image captions},
 author={Singla, Vasu and Yue, Kaiyu and Paul, Sukriti and Shirkavand, Reza and Jayawardhana, Mayuka and Ganjdanesh, Alireza and Huang, Heng and Bhatele, Abhinav and Somepalli, Gowthami and Goldstein, Tom},
 journal={arXiv preprint arXiv:2406.10328},
 year={2024}
}

@article{beyer2024paligemma,
 title={Paligemma: A versatile 3b vlm for transfer},
 author={Beyer, Lucas and Steiner, Andreas and Pinto, Andr{\'e} Susano and Kolesnikov, Alexander and Wang, Xiao and Salz, Daniel and Neumann, Maxim and Alabdulmohsin, Ibrahim and Tschannen, Michael and Bugliarello, Emanuele and others},
 journal={arXiv preprint arXiv:2407.07726},
 year={2024}
}

@misc{nostalgebraist2020logitlens,
 author = {nostalgebraist},
 title = {interpreting gpt: the logit lens},
 year = {2020},
 howpublished = {\url{https://www.lesswrong.com/posts/AcKRB8wDpdaN6v6ru/interpreting-gpt-the-logit-lens}}
}

@article{krojer2026latentlens,
 title={LatentLens: Revealing Highly Interpretable Visual Tokens in LLMs},
 author={Krojer, Benno and Nayak, Shravan and Mañas, Oscar and Adlakha, Vaibhav and Elliott, Desmond and Reddy, Siva and Mosbach, Marius},
 journal={arXiv preprint arXiv:2602.00462},
 year={2026}
}

\appendix

\section{Other Models}\label{sec:other-models}
For the purposes of output replication, we repeat these experiments on other models in the Appendix. We use \textsc{Qwen2-VL-2B-Instruct}, a pretrained model by Alibaba. Experiments with this model dynamically process images with a patch size of 14x14 pixels, while text input is tokenized systematically through byte pair encoding based on tiktoken \citep{shantanu2025tik}. The model uses a \textsc{Qwen2}-specific Vision Transformer Architecture that is optimized by SwiGLU, which affects activation function \citep{shazeer2020glu}, and RMSNorm, which replaces the traditional layer normalization technique \citep{jiang2023pre, bai2023qwen}. This is implemented throughout 28 layers.

Additionally, we perform experiments on \textsc{PaliGemma2-10B-pt-448}. Similarly to \textsc{PaliGemma-3B-pt-448}, it uses a SigLIP encoder, but upgrades its text decoder to \textsc{Gemma2}. \citep{keysers2024introducing} This model is used to ensure scalability by increasing \textsc{PaliGemma}'s parameters.

\section{Attention vs MLP Contributions}\label{sec:attention-vs-mlp-contributions}

\begin{figure}[htbp]
\centering
\begin{subfigure}{.48\linewidth}
  \centering
  \includegraphics[width=\linewidth]{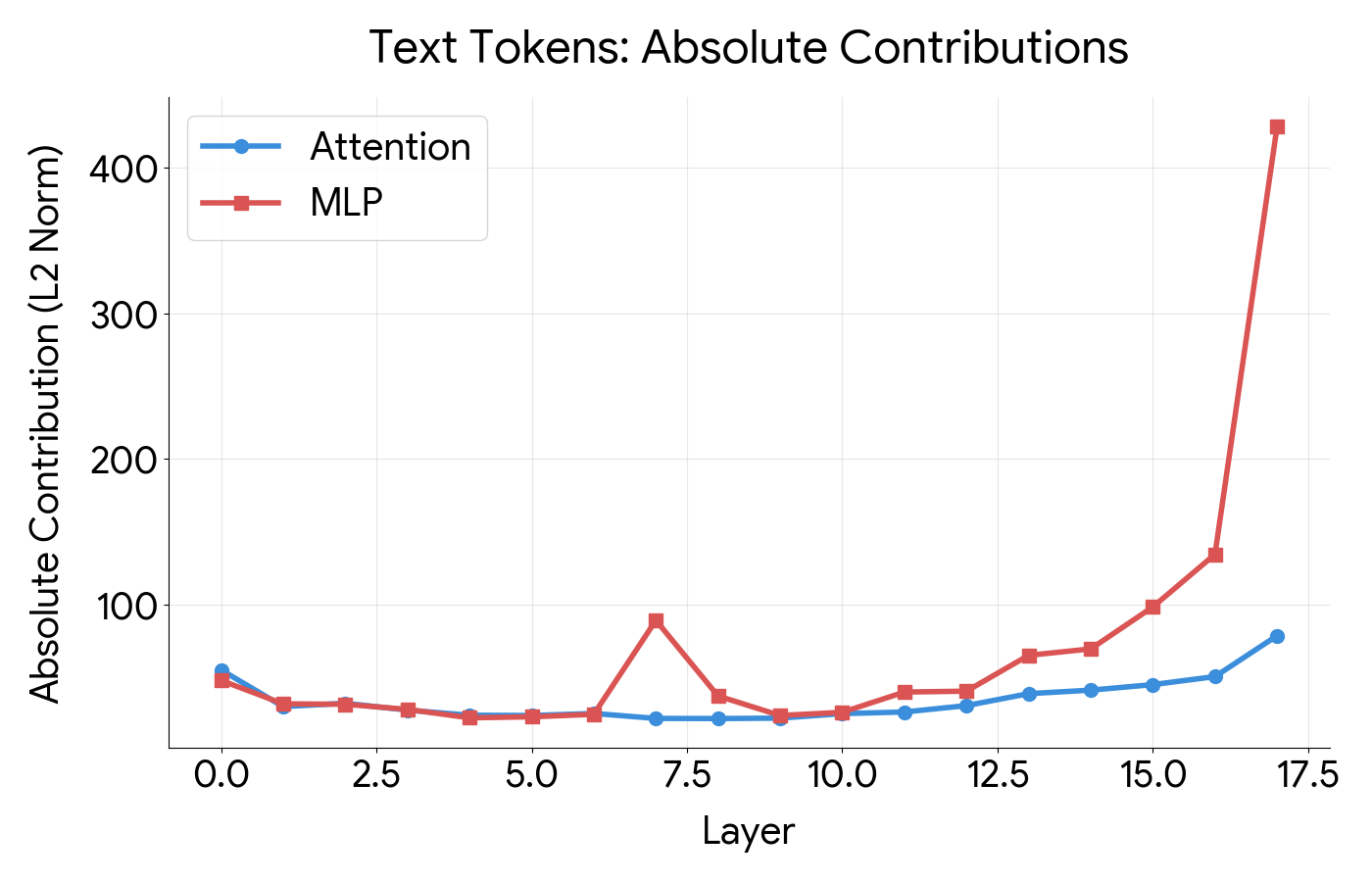}
  \caption{Text Tokens Absolute Contributions}
  \label{fig:sub3}
\end{subfigure}\hfill
\begin{subfigure}{.48\linewidth}
  \centering
  \includegraphics[width=\linewidth]{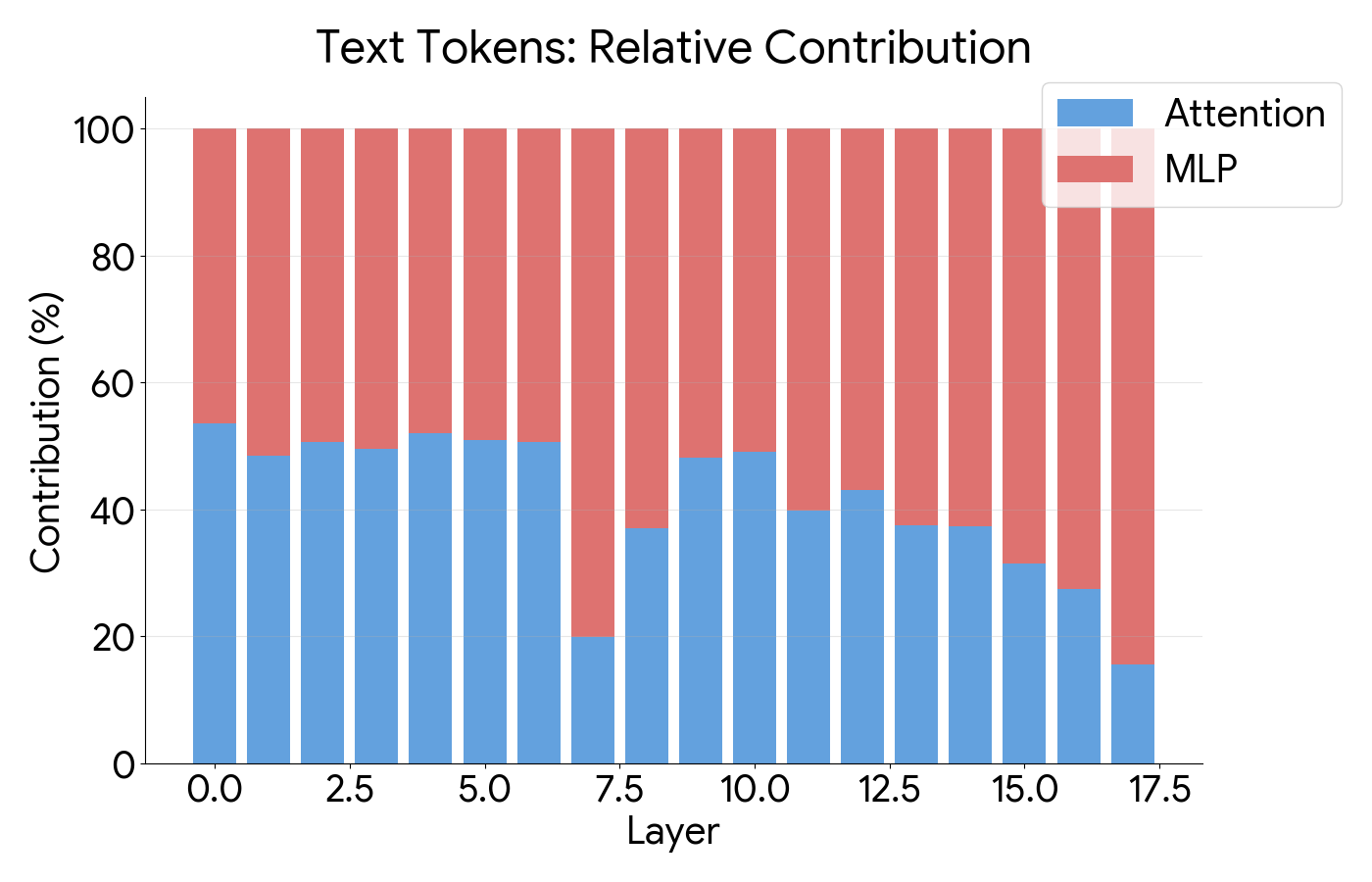}
  \caption{Text Tokens Relative Contributions}
  \label{fig:sub4}
\end{subfigure}
\caption{Attention and MLP contributions to text token updates: Relative contributions reveal attention and MLP contribute roughly equally throughout most layers, with attention comprising 40-60\% of total changes. Absolute contributions shows that MLP spikes at Layer 7 to an L2 Norm of 100 before stabilizing, then surges once more at the final layer to 400. }
\label{fig:attention-vs-mlp-contributions-to-text-tokens}
\end{figure}

We analyze the attention and MLP sublayers for text tokens. Figure \ref{fig:attention-vs-mlp-contributions-to-text-tokens} demonstrates that between these two residual stream contributions, attention consistently increases between layers while MLP spikes dramatically in late layers. However, compared to the absolute total contributions in Figure \ref{fig:attention-vs-mlp-contributions-to-vision-tokens}, the contribution is much lower.

\subsection{Qwen2}
\begin{figure}[htbp]
\centering
\begin{subfigure}{.48\linewidth}
  \centering
  \includegraphics[width=\linewidth]{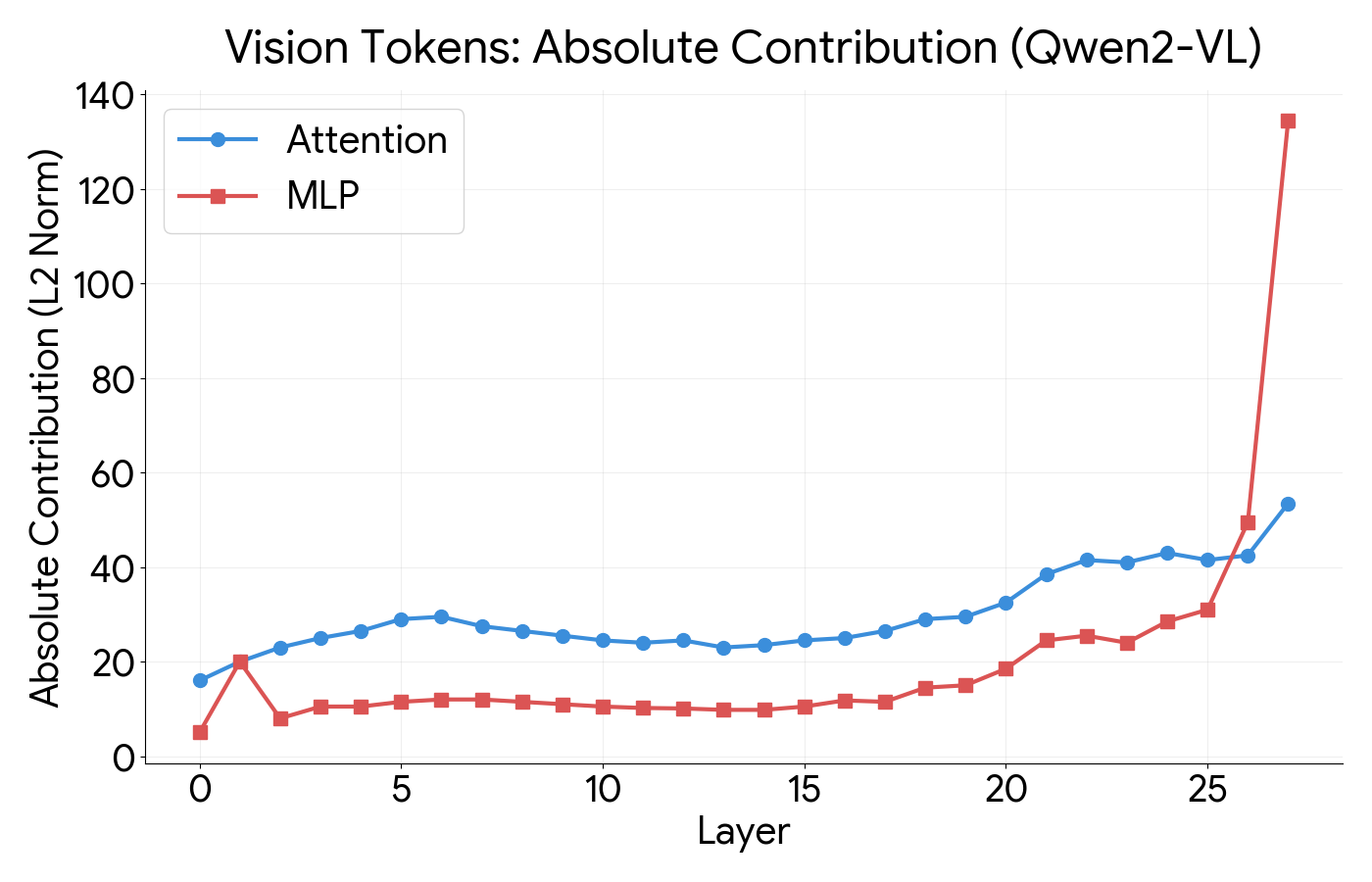}
  \caption{Vision Token Absolute Contributions}
  \label{fig:sub5}
\end{subfigure}\hfill
\begin{subfigure}{.48\linewidth}
  \centering
  \includegraphics[width=\linewidth]{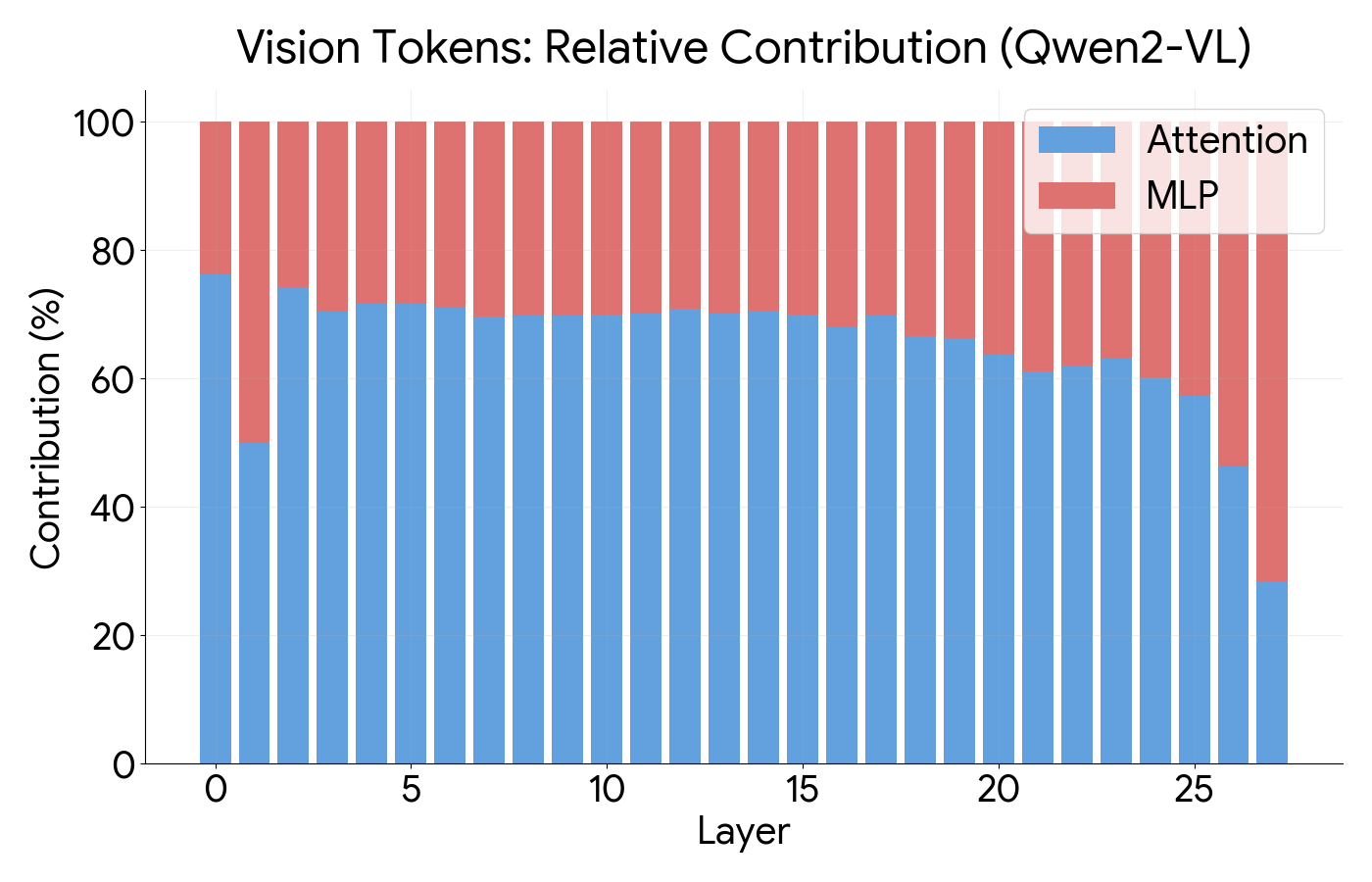}
  \caption{Vision Token Relative Contributions}
  \label{fig:sub6}
\end{subfigure}
\caption{\textsc{Qwen2}'s Attention and MLP contributions to vision token updates: Relative contributions reveal attention consistently contributes more than MLP to the residual stream in early to intermediate layers, with a small spike in MLP during the very early layers. In late layers, MLP drastically contributes to the residual stream.}
\label{fig:qwen-attention-mlp-vision-tokens}
\end{figure}

\begin{figure}[htbp]
\centering
\begin{subfigure}{.48\linewidth}
  \centering
  \includegraphics[width=\linewidth]{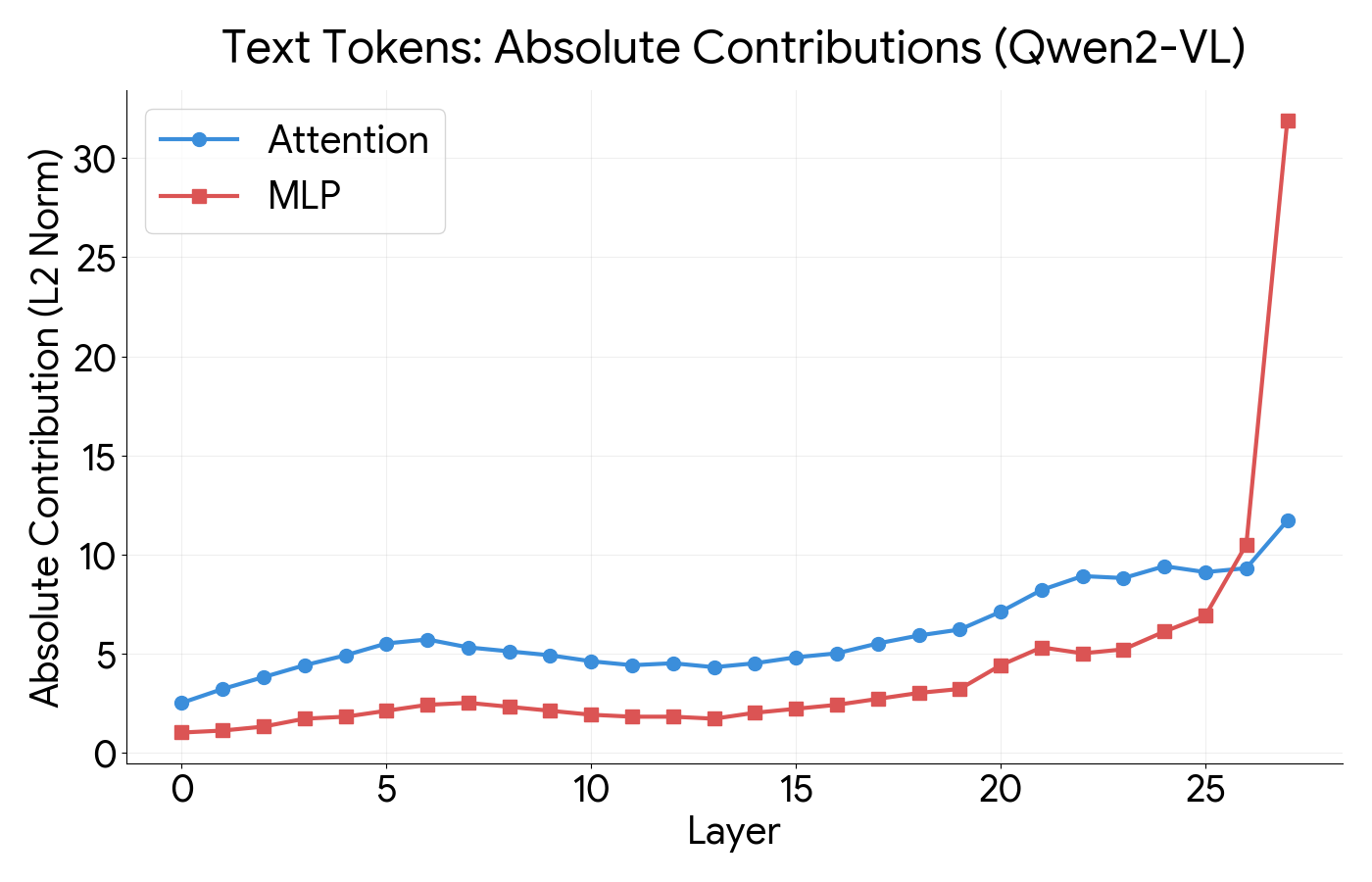}
  \caption{Text Token Absolute Contributions}
  \label{fig:sub7}
\end{subfigure}\hfill
\begin{subfigure}{.48\linewidth}
  \centering
  \includegraphics[width=\linewidth]{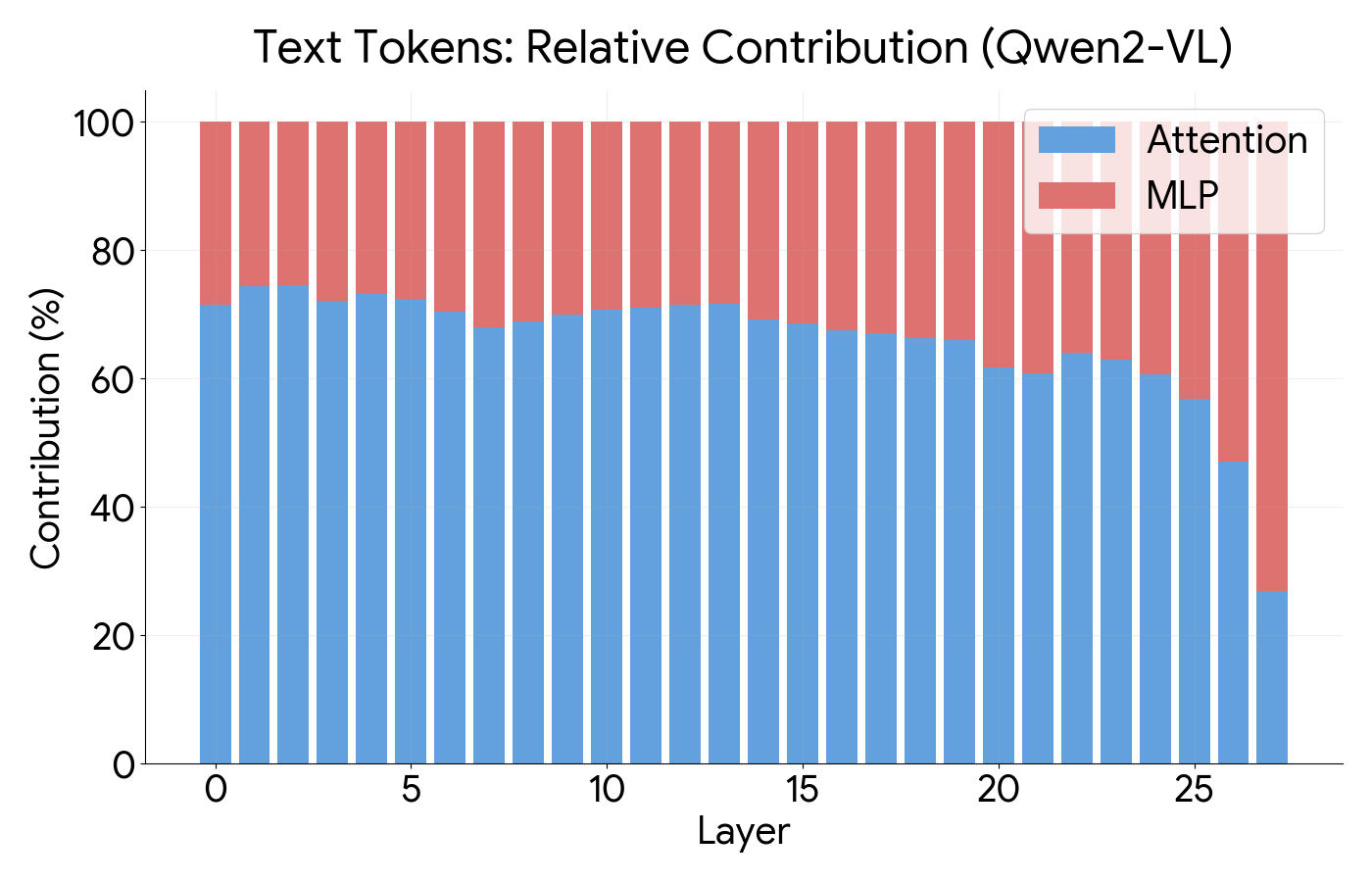}
  \caption{Text Token Relative Contributions}
  \label{fig:sub8}
\end{subfigure}
\caption{\textsc{Qwen2}'s Attention and MLP contributions to text token updates: Relative contributions reveal attention consistently contributes more than MLP to the residual stream in early to intermediate layers. In late layers, MLP drastically contributes to the residual stream.}
\label{fig:qwen-attention-mlp-text-tokens}
\end{figure}

Next, we replicate these experiments in \textsc{Qwen2-VL-2B-Instruct}. In both Figure \ref{fig:qwen-attention-mlp-vision-tokens} and Figure \ref{fig:qwen-attention-mlp-text-tokens}, attention is the major contributor to the residual stream. This aligns with \textsc{PaliGemma}'s distribution whose main contributions are sourced from attention. However, attention in \textsc{PaliGemma} only overtakes MLP contribution by a slight amount, while the distribution for \textsc{Qwen2} shows a much larger degree of favorability towards attention.

In accordance to vision tokens, MLP contribution in \textsc{Qwen2} has a small spike in the very early layers and a large spike in late layers. This shows when MLP processes visual features: in very early layers, the raw data is marked. During the late layers, MLP finally transforms the data during cross-modal integration. This differs from Figure \ref{fig:attention-vs-mlp-contributions-to-text-tokens} in that \textsc{PaliGemma} spikes in MLP in early-to-intermediate layers, rather than in the second layer such as \textsc{Qwen2}. 

In Figure \ref{fig:qwen-attention-mlp-text-tokens}, MLP's contribution to the residual stream spikes in late layers. These results are similar to the ones found in \textsc{PaliGemma}'s contributions: the textual features are being integrated alongside visual features for the final output. However, in contrast to \textsc{PaliGemma}'s results, attention contributes to a higher degree overall at nearly 60\%-70\% until the late layers.

\subsection{PaliGemma2-10B}
Finally, we determine relative and absolute contributions on a \textsc{PaliGemma} model scaled to 10B parameters. Despite using the same vision encoder, the attention and MLP contributions shown in the vision token results are abnormal compared to the results seen in \textsc{PaliGemma-3B} and \textsc{Qwen2}. Both MLP and attention contributions contribute at high rates in early layers, with MLP being the priority contributor. However, in subsequent layers, attention becomes the majority contributor. This is in contrast to the 2B and 3B parameter models which start off low in contributions to the residual stream, with a small rise in attention and a large spike in MLP at the very last layer. 

\begin{figure}[htbp]
\centering
\begin{subfigure}{.48\linewidth}
  \centering
  \includegraphics[width=\linewidth]{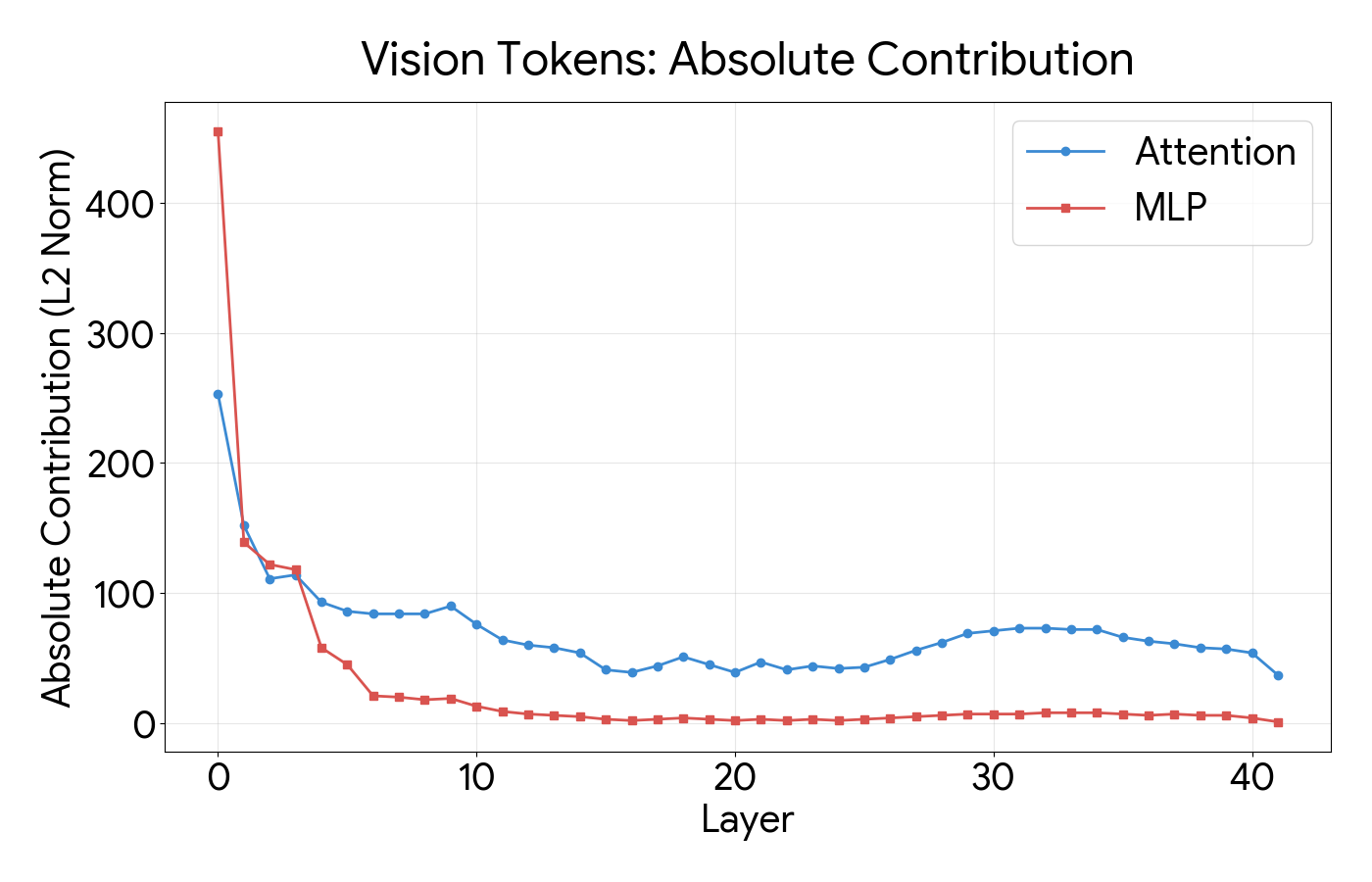}
  \caption{Vision Token Absolute Contributions}
  \label{fig:sub9}
\end{subfigure}\hfill
\begin{subfigure}{.48\linewidth}
  \centering
  \includegraphics[width=\linewidth]{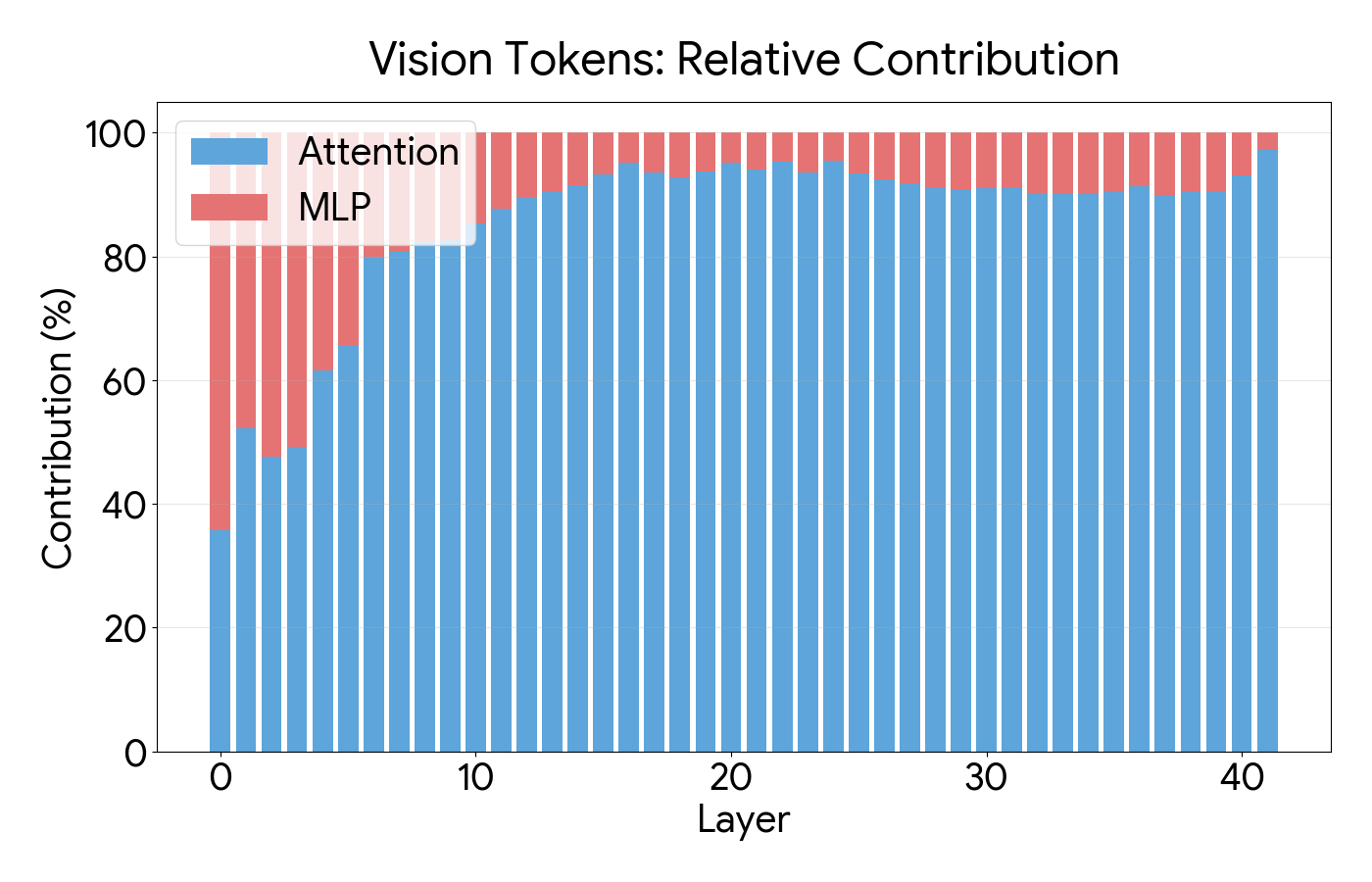}
  \caption{Vision Token Relative Contributions}
  \label{fig:sub10}
\end{subfigure}
\caption{\textsc{PaliGemma2-10B}'s Attention and MLP contributions to vision token updates: Relative contributions reveal attention consistently contributes more than MLP to the residual stream in intermediate to late layers. In early layers, MLP drastically contributes to the residual stream.}
\end{figure}

\begin{figure}[htbp]
\centering
\begin{subfigure}{.48\linewidth}
  \centering
  \includegraphics[width=\linewidth]{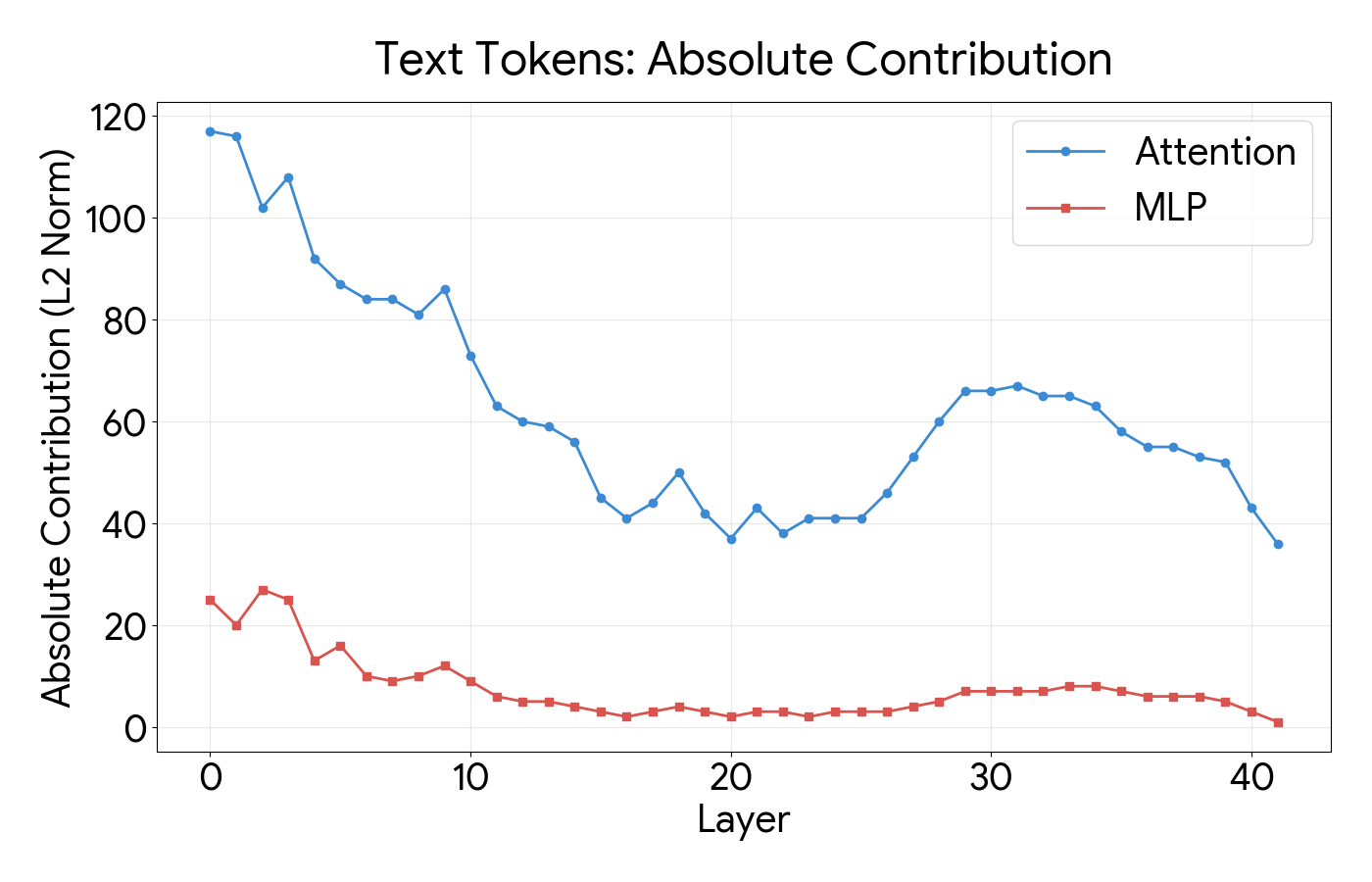}
  \caption{Text Token Absolute Contributions}
  \label{fig:sub11}
\end{subfigure}\hfill
\begin{subfigure}{.48\linewidth}
  \centering
  \includegraphics[width=\linewidth]{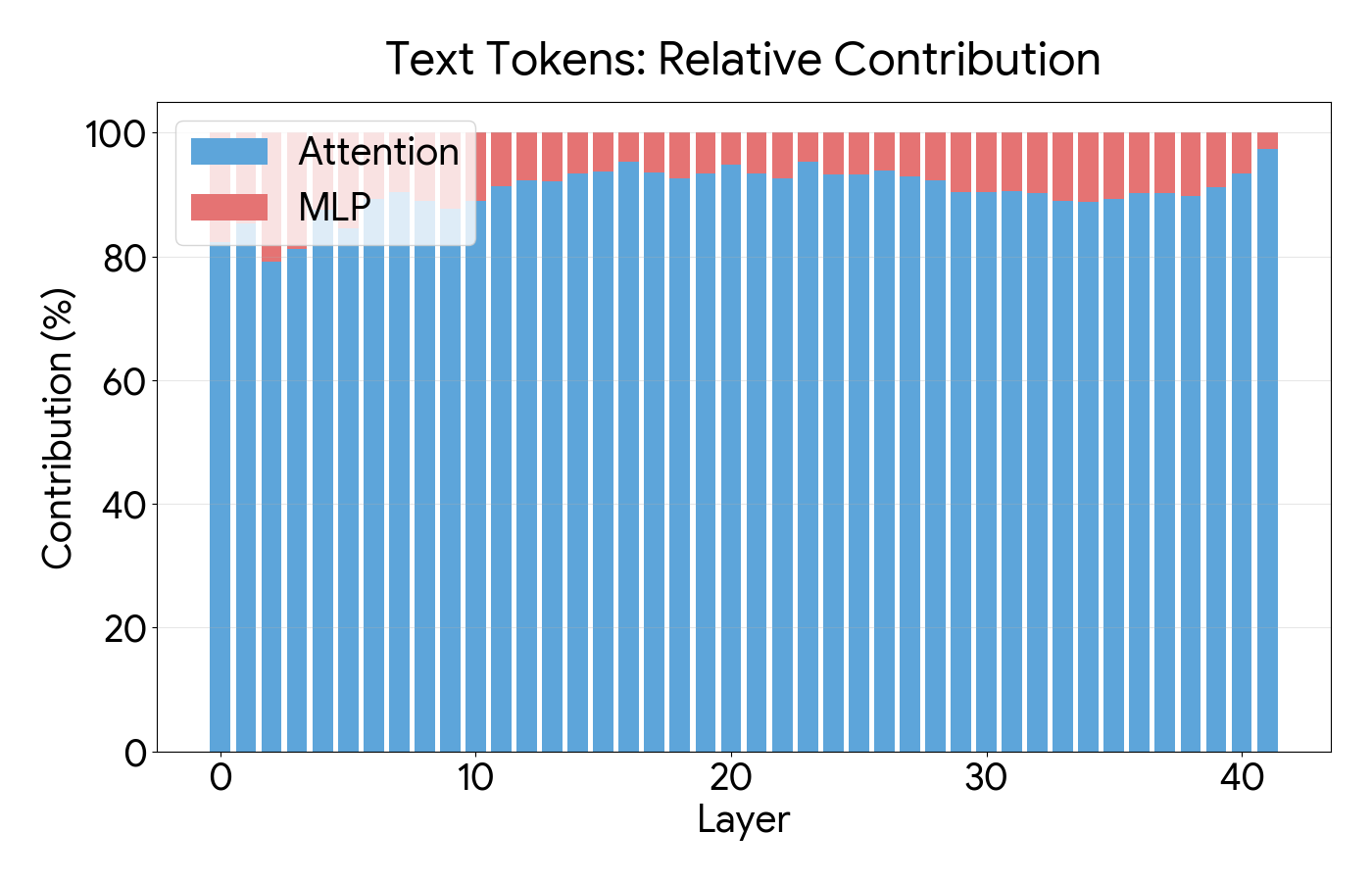}
  \caption{Text Token Relative Contributions}
  \label{fig:sub12}
\end{subfigure}
\caption{\textsc{PaliGemma2-10B}'s Attention and MLP contributions to text token updates: Relative contributions reveal attention contributes more than MLP to the residual stream consistently. Attention spikes in layers 25-30.}
\end{figure}

The text tokens result in similar behavior in that contributions start high and diminish overtime. However, attention spikes at the intermediate layers. Distribution-wise, attention takes over 90\% of the contributions to the residual stream consistently. Unlike the other models which spiked in MLP contributions at late layers, \textsc{PaliGemma2-10B}'s distribution spiked in MLP contributions at the early layers.

\section{Attention Contributions Visualized}\label{sec:attentino-contributions-visualized}

\subsection{Qwen2}

\begin{figure}[htbp]
   \centering
   \includegraphics[width=\linewidth]{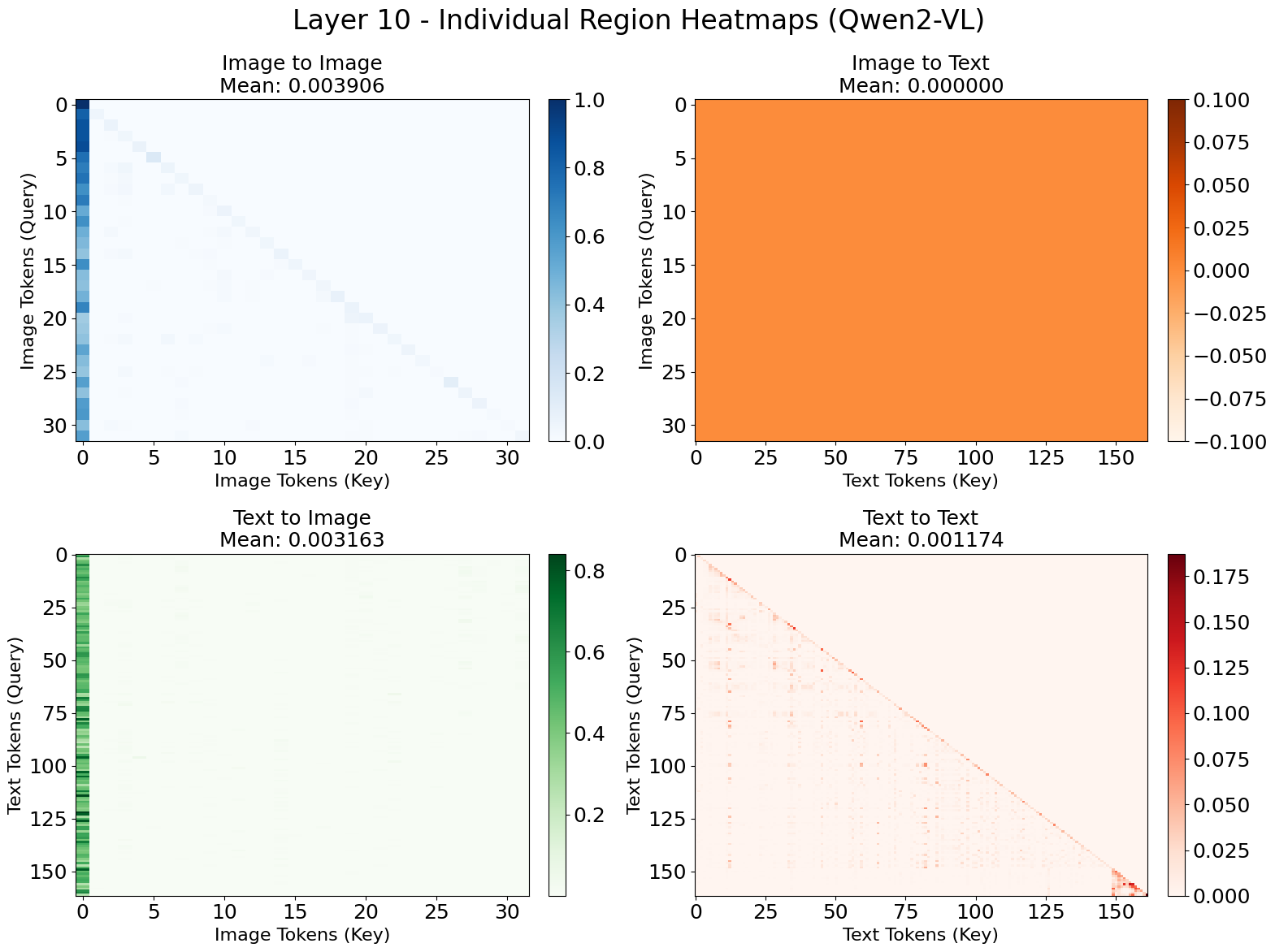}
   \caption{Attention score decomposition for \textsc{Qwen2-VL-2B}: Each quadrant shows attention from query tokens (rows) to key tokens (columns). Top-left: Vision queries attending to vision keys (V to V, mean=0.003906). Top-right: Vision queries attending to text keys (V to T, mean=0). Bottom-left: Text queries attending to vision keys (T to V, mean=0.003163). Bottom-right: Text queries attending to text keys (T to T, mean=0.001174). }
   \label{fig:qwen-attention-scores-sub-contributions2}
\end{figure}

Similar to the experiment performed in \textsc{PaliGemma}, Figure \ref{fig:qwen-attention-scores-sub-contributions2} demonstrates that each query token primarily attends to a key of the same modality. As further evidence, the image token queries do not attend to the text token keys at all, indicating the lack of cross-modal integration towards text token keys. This signals a structural bias, in which both vision and text queries attend to image keys while only text queries attend to text keys. 

Broadly, this suggests that \textsc{Qwen2} has a textual bias, allowing cross-modal integration to occur in only one direction. On a lesser note, the image queries attend to the image keys more than the text queries do. However, the mean scores still lie within a similar range of each other. The relatively equal distribution, in which both modalities are emphasized when attending to image token keys, further signals model reliance on text information for visual output.

\subsection{PaliGemma2-10B}

Next, we test how query attention distributes to keys of different modalities in a scaled model. Since it is part of the \textsc{PaliGemma} family, we expect to see the results will follow an attention trend similar to Figure \ref{fig:attention-scores-sub-contributions}. 

\begin{figure}[htbp]
   \centering
   \includegraphics[width=\linewidth]{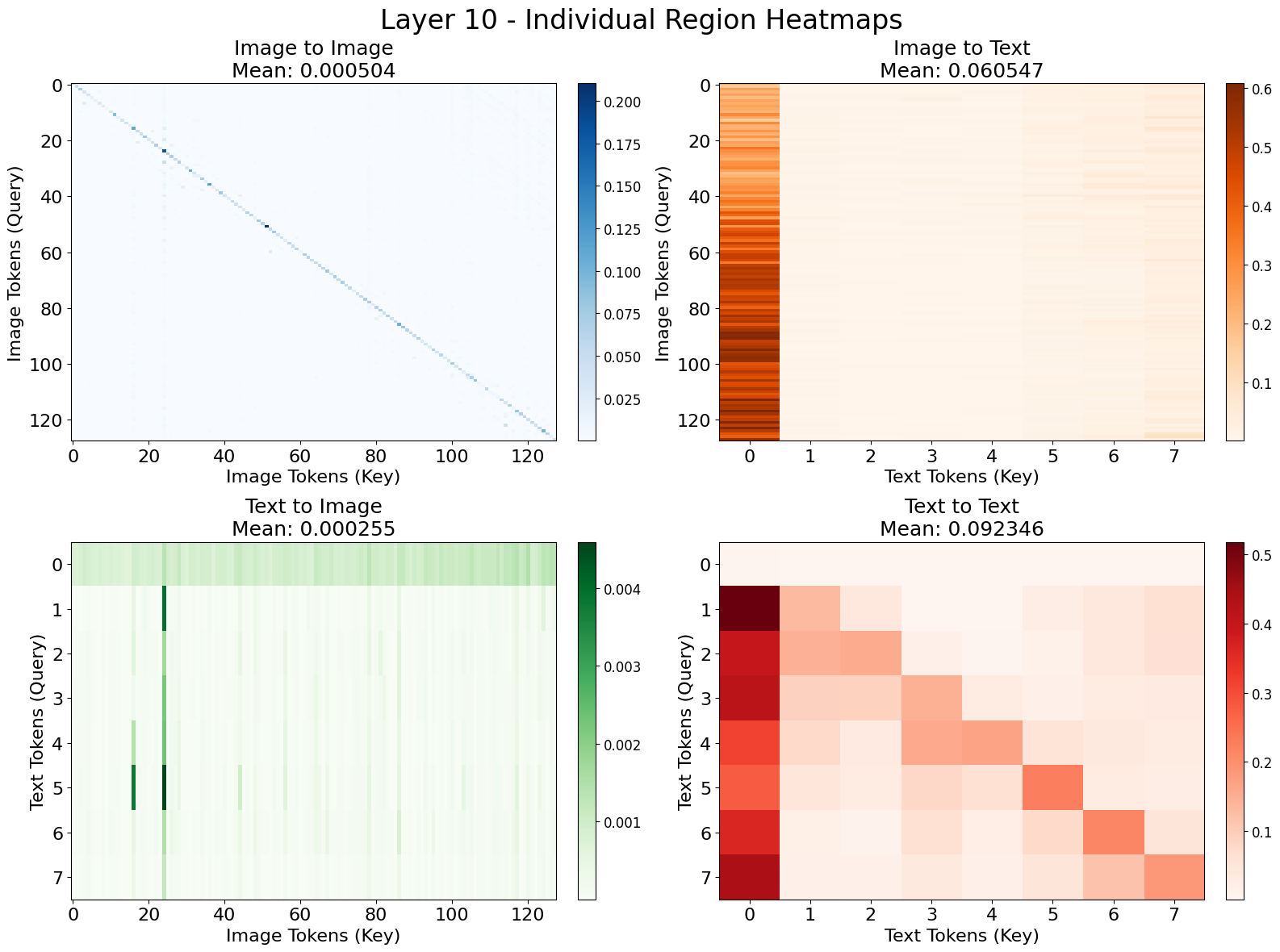}
   \Description{Attention score decomposition for \textsc{PaliGemma2-10B}: Each quadrant shows attention from query tokens (rows) to key tokens (columns). Top-left: Vision queries attending to vision keys (V to V, mean=0.000504). Top-right: Vision queries attending to text keys (V to T, mean=0.060547). Bottom-left: Text queries attending to vision keys (T to V, mean=0.000255). Bottom-right: Text queries attending to text keys (T to T, mean=0.092346).}
   \caption{Attention score decomposition for \textsc{PaliGemma2-10B}: Each quadrant shows attention from query tokens (rows) to key tokens (columns). Top-left: Vision queries attending to vision keys (V to V, mean=0.000504). Top-right: Vision queries attending to text keys (V to T, mean=0.060547). Bottom-left: Text queries attending to vision keys (T to V, mean=0.000255). Bottom-right: Text queries attending to text keys (T to T, mean=0.092346). }
   \label{fig:paligemma10b-attention-scores-sub-contributions}
\end{figure}

As expected, image token keys are derive most of its attention from image queries rather than the text queries, and vice versa for text token keys. However, it is notable that text keys are attended to at large rates by both image and text queries. Image queries have a mean attention score that is greater than half, while the score as attended to by text queries is greater than 0.90.

\section{Activation Front Patching Experiments}\label{sec:full-layer-patching}

\subsection{Qwen2-VL}\label{sec:qwen-text-patching}
\subsubsection{Experimental Setup}
Within the \textsc{Qwen2-VL-2B-Instruct} architecture, we perform Front Patching, similar to \textsc{PaliGemma}, by extracting internal activation states from earlier layers and injecting them into the final three layers of the model. 

\subsubsection{Textual Representation Patching Results}
By patching different configurations of layers (2-4, 5-7, 7-9, and 10-12) into the last three layers, the language model of \textsc{Qwen2-VL} severely collapsed. Across all configurations, the logit score for the correct answer dropped from a highly confident score of 15.79 to around 1.0-2.3. The LLM's residual stream experiences don't benefit from patching and require linear transformation. The extreme logit drop reveals that the "concept" of the target logit "2" is not retrieved immediately. The model likely spends the intermediate layers integrating the visual tokens with the text prompt, which would've increased the probability of the counting token. Patching across these configurations underscores the importance of the reasoning phase, where both modalities align.

\subsubsection{Vision Representation Patching Results}
Patching earlier layers of the vision tower into the final layer improved the logit score by approximately 0.65, which is a slight performance increase. However, for other source layer configurations, there were negligible performance drops, maintaining the model's confidence in the correct answer. This suggests that the fundamental features required for "counting" (what the model was specifically prompted with) are established in the very early layers of the encoder. 

\begin{table}[htbp]
\centering
\footnotesize 
\setlength{\tabcolsep}{5pt} 
\caption{Layer-wise Representation Patching Results for Qwen2-VL, patching source layer activations into late target layers reveals a stark modality asymmetry in \textsc{Qwen2-VL}. While visual features remain highly stable (showing slight performance improvements and stable loss), early textual representations are fundamentally in conflict with late layers. The resulting catastrophic output collapse, marked by logit drops up to -14.77, confirms that the progressive processing requirement for text tokens generalizes across different adapter-based architectures.}
\resizebox{\linewidth}{!}{%
\begin{tabular}{lcccccc}
\hline
\textbf{Modality} & \textbf{Layers} & \textbf{Clean} & \textbf{Patched} & \textbf{$\Delta$ Logit} & \textbf{Clean Loss} & \textbf{Patched Loss} \\
\hline
Vision & 2--4   & 15.78 & 16.42 & +0.65 & 19.58 & 20.18 \\
Vision & 5--7   & 15.78 & 15.49 & -0.29 & 19.58 & 20.33 \\
Vision & 7--9   & 15.79 & 15.40 & -0.39 & 19.57 & 20.20 \\
Vision & 10--12 & 15.77 & 15.49 & -0.28 & 19.58 & 20.60 \\
\hline
Text & 2--4   & 15.79 & 2.03  & -13.76 & 19.57 & 12.09 \\
Text & 5--7   & 15.79 & 1.82  & -13.97 & 19.57 & 10.09 \\
Text & 7--9   & 15.79 & 1.01  & -14.77 & 19.57 & 11.24 \\
Text & 10--12 & 15.79 & 2.29  & -13.50 & 19.57 & 12.00 \\
\hline
\end{tabular}%
}
\end{table}

\subsection{PaliGemma-10B}
We conduct residual stream activation patching on 82 valid image samples in \textsc{paligemma2-10b-pt-448} to measure the semantic stability of visual and textual tokens throughout the model's 42 layers. We perform a front patching experiments from layers 2-6 and layers 18-22 onto the final layers, 38-41.

\subsubsection{Textual Representation Patching Results}\label{sec:paligemma2b-text-front}
The text stream was revealed to be extremely sensitive, as shown by the drop in the $\Delta$ logit from -16.70 when patching from early to late layers. The textual representations in earlier layers are incompatible with those in later layers, indicating that the model requires the full length of the network to fully "process" the text tokens needed for answer generation.

\subsubsection{Vision Representation Patching Results}
In stark comparison, the vision tokens remain linearly stable throughout the network. Patching visual states from early or mid layers into later layers resulted in a similar performance improvement with a +3.51-+3.59 $\Delta$ Logit. This indicates that early and mid layer visual representations are already semantically usable for the final classification layer. Especially for the earlier layers, this implies that skipping over the intermediate processing layers may provide the output heads with a better visual signal. 

\begin{table}[htbp]
\centering
\footnotesize 
\setlength{\tabcolsep}{5pt} 
\caption{Layer-wise Representation Patching Results for PaliGemma2-10B: To evaluate whether our findings scale, we replicated the activation patching experiment on the 10B parameter model. The results strongly mirror \textsc{PaliGemma-3B}: patching vision representations from early or mid layers into later layers actually improves performance ($\Delta$ Logit +3.59), indicating that visual states are semantically usable early on. Conversely, patching early text representations leads to massive distributional collapse (loss spikes to 76.71).}
\label{tab:paligemma2_patching}
\resizebox{\linewidth}{!}{
\begin{tabular}{lcccccc}
\hline
\textbf{Modality} & \textbf{Layers} & \textbf{Clean} & \textbf{Patched} & \textbf{$\Delta$ Logit} & \textbf{Clean Loss} & \textbf{Patched Loss} \\
\hline
Vision & 2--6 & 12.84 & 16.35 & +3.51 & 12.26 & 9.77 \\
Vision & 18--22 & 12.85 & 16.44 & +3.59 & 12.26 & 10.03 \\
\hline
Text & 2--6 & 12.84 & -3.86 & -16.70 & 12.26 & 76.71 \\
Text & 18--22 & 12.84 & 12.63 & -0.21 & 12.26 & 11.64 \\
\hline
\end{tabular}%
}
\end{table}

\section{Causal Tracing Front Patching}

\subsection{Qwen2-VL-2B}

In \textsc{Qwen2-VL}, the visual representations appear virtually identical across the entire depth of the network. When patching early visual tokens directly into the final layers, it yielded an output distribution nearly identical to the baseline ($\mu = 0.0094, \sigma = 0.0184$). This demonstrates that post-projection visual tokens are fully processed, static representations that can bypass the intermediate layers. Conversely, the text pathway exhibits standard progressive processing requirements, as seen when patching early text tokens into late layers, which led to a severe distributional collapse  ($\mu = 11.3625, \sigma = 0.3034$).
\begin{figure}[htbp]
    \centering
    \includegraphics[width=\linewidth, height=5cm]{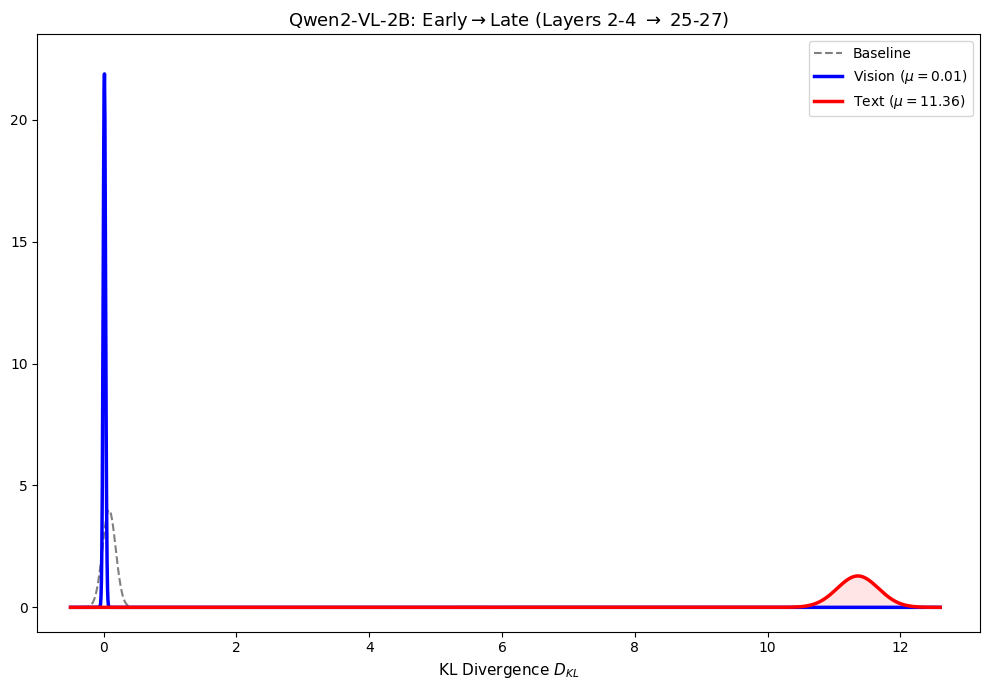}
    \Description{Output Distribution Shifts Measured via KL Divergence for \textsc{Qwen2-VL-2B}: The density plot compares the natural baseline output variation against interventions where early layer activations (2--4) are patched into late layers (25--27). When visual activations are patched from early to late layers, the resulting distribution shift ($\mu=0.01$) is nearly identical to the baseline. Conversely, patching early text activations into late layers results in a severe distributional collapse ($\mu=11.36$).}
    \caption{Output Distribution Shifts Measured via KL Divergence for \textsc{Qwen2-VL-2B}: The density plot compares the natural baseline output variation against interventions where early layer activations (2--4) are patched into late layers (25--27). When visual activations are patched from early to late layers, the resulting distribution shift ($\mu=0.01$) is nearly identical to the baseline. Conversely, patching early text activations into late layers results in a severe distributional collapse ($\mu=11.36$).}
    \label{fig:qwenkl}
\end{figure} 

\subsection{PaliGemma2-10B}

When early vision representations (layers 2–4) were patched directly into the late layers (layers 39–41), the model produced a near-baseline KL divergence ($\mu = 0.28, \sigma = 0.19$). This near-zero divergence indicates that vision representations don't change across the depth of the language model; once projected through the adapter, they act semantically stable. The textual pathway yields opposite results, validating the idea that it depends on sequential processing. Patching early text activations into the late layers resulted in a catastrophic distributional collapse ($\mu = 40.89, \sigma = 5.36$), proving that text tokens at Layer 0 lack the density required to support late-stage reasoning without intermediate processing. 

\begin{figure}[htbp]
    \centering
    \includegraphics[width=\linewidth, height=5cm]{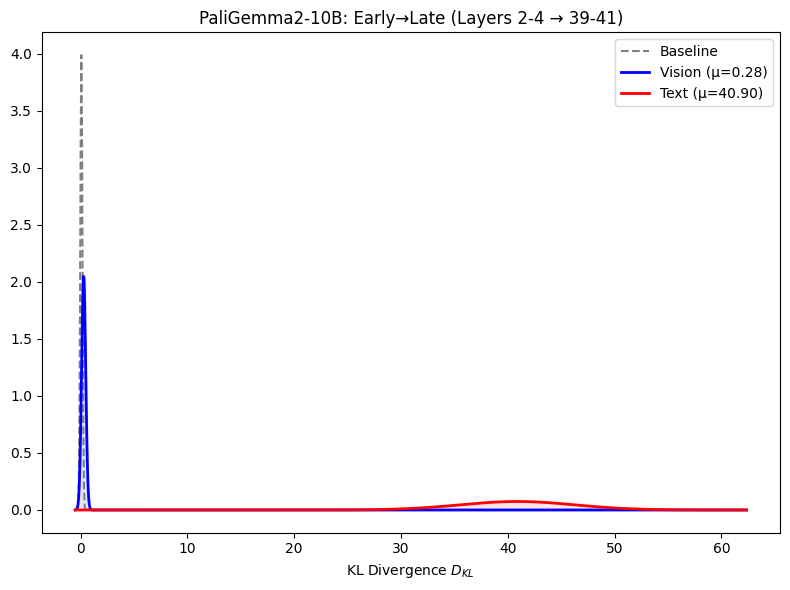}
    \Description{Output Distribution Shifts Measured via KL Divergence for \textsc{PaliGemma2-10B}: The density plot compares the natural baseline output variation against interventions where early layer activations (2--4) are patched into late layers (39--41). When visual activations are patched from early to late layers, the model produces a near-zero distribution shift ($\mu=0.28$), indicating high semantic stability across depth. Conversely, patching early text activations into the late layers results in a catastrophic distributional collapse ($\mu=40.90$).}
    \caption{Output Distribution Shifts Measured via KL Divergence for \textsc{PaliGemma2-10B}: The density plot compares the natural baseline output variation against interventions where early layer activations (2--4) are patched into late layers (39--41). When visual activations are patched from early to late layers, the model produces a near-zero distribution shift ($\mu=0.28$), indicating high semantic stability across depth. Conversely, patching early text activations into the late layers results in a catastrophic distributional collapse ($\mu=40.90$).}
    \label{fig:paligemma2-10bkl}
\end{figure} 

\section{kNN Analysis on Text and Vision}

\subsection{Qwen2-VL-2B}

The kNN probe of \textsc{Qwen2-VL's} Layer 0 reveals an embedding matrix defined by multilingual lexical parity rather than conceptual depth. Within the semantic boundary, the model largely fails to retrieve true English synonyms (e.g., "truck" or "transport"), instead organizing its neighbors around repetitive script-based lexical variations and transliterations. After crossing the threshold, the vector space begins to collapse into statistical noise, where the model pairs target concepts with fractured substrings (e.g., 'phil', 'cean') and loosely related multilingual fragments that share nearly identical cosine similarity scores. This degraded space exhibits a distinct distributional bias by linking conceptually distinct entities into the same narrow similarity band as the target. These findings reinforce that \textsc{Qwen2-VL}'s text tokens are raw at entry.

For \textsc{Qwen2-VL}, the vision-token experiment is adapted to reflect its different visual architecture. Unlike \textsc{PaliGemma}, which uses a SigLIP vision tower followed by a separate projector, \textsc{Qwen2-VL} first embeds raw image patches, processes them through vision transformer blocks, and then applies a learned patch merger that combines local 2x2 patch groups into language-model-ready visual tokens. Therefore, the merged patch layer is compared against the vision-blocked patches, rather than using the SigLIP/post-projection split used for \textsc{PaliGemma-3B} and \textsc{PaliGemma2-10B}.

In the merged patch layer, the two anchors have low cross-anchor similarity. However, the similarity of the nearest neighbors for the blueberry patch ranges between 0.25-0.45, while the strawberry patch ranges between 0.75-0.80. Their nearest neighbors seem to be more spatially and visually scattered without having gone through contextualization. Rather, the patches seem to be compared using color. The initial blueberry patch is dark with a shine on one side, which \textsc{Qwen2} may be picking out in relation to the patches of white plate with a dark shadow underneath. Since this is not completely accurate to the coloring of the initial patch, the blueberry's cosine similarity is much smaller than the strawberry patch's similarity. This differs from the \textsc{PaliGemma} models, which choose the nearest neighboring patches as the most semantically relevant. 

After the vision blocks and merger, the cross-anchor similarity increases, suggesting that the visual tokens become more context-aware and are partially organized around shared scene-level information. The neighbors identified in vision-blocked patches reflect the anchor tokens, similar to \textsc{PaliGemma}.

\clearpage
\subsection{PaliGemma2-10B}

While \textsc{PaliGemma2-10B}'s semantic boundary successfully retrieves high-level synonyms like 'vehicle' and 'automobile', it quickly reverts to organizing neighbors around lexical variations like 'cat' and 'bar'. Immediately upon crossing the semantic boundary, the vector space undergoes a total collapse into statistical noise, where the model pairs the target concept 'car' with completely unrelated tokens, such as 'man', 'house', and 'ch', that occupy the same similarity value. This space reflects a severe distributional bias, where character-level overlap takes precedence over actual semantic meaning. This evidence demonstrates that uncontextualized text tokens are semantically raw and contain phonetic and structural confusion.

Next, the same vision token experiment is tested for \textsc{PaliGemma2-10B}. Immediately, it can be seen that the images are split into smaller patches, therefore having a larger number of patches than \textsc{PaliGemma-3B}. Similar to the vision token results for \ref{fig:vision-knn}, the SigLIP layer 0 neighbors remain visually compact around the selected fruit patches, where nearby fruit patches are retrieved as neighbors. Additionally, post-projection, the neighbors become semantically meaningful and visually reflect the anchors within the image.

\section{Residual Stream Updates}
\label{sec:rs-updates}
To determine if the multimodal representation dynamics observed in \textsc{PaliGemma-3B} generalize across different VLMs, we replicated our residual stream tracking experiment on \textsc{Qwen2-VL} and \textsc{PaliGemma2-10B}.

\begin{figure}[htbp]
\centering
\begin{minipage}{0.48\textwidth}
  \centering
  \includegraphics[width=\linewidth]{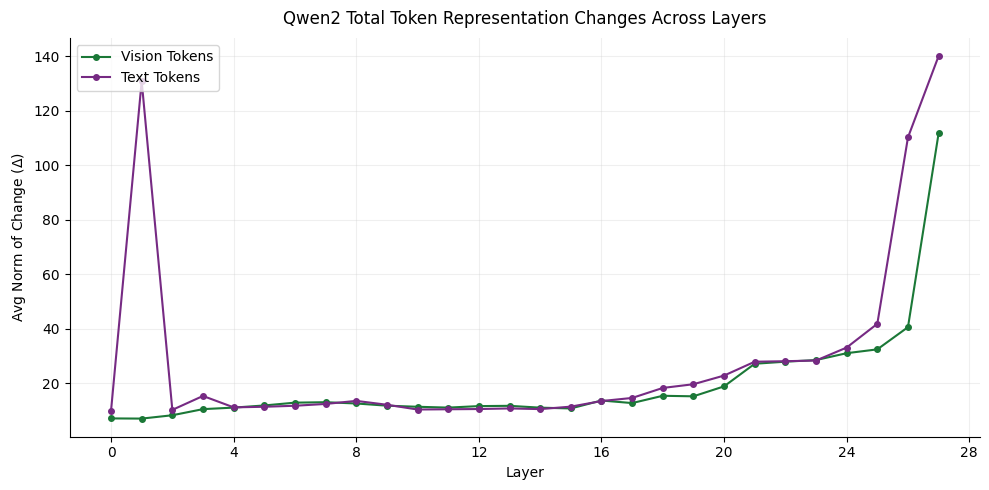}
  \Description{\textbf{Layer-wise residual stream update norms for Qwen2-VL.} The plot illustrates the average norm of change ($\Delta$) for both vision and text token representations across the model's layers. A distinctive, massive initial spike in text token updates occurs specifically at Layer 1. This is followed by a prolonged, stable fusion phase (from Layers 4--20), during which both modalities exhibit low and consistent representation changes. In the final layers (21--28), a significant escalation is observed, resulting in sharp increases prior to the final output, with text token updates peaking at a higher magnitude than vision tokens.}
  \caption{\textbf{Layer-wise residual stream update norms for Qwen2-VL.} The plot illustrates the average norm of change ($\Delta$) for both vision and text token representations across the model's layers. A distinctive, massive initial spike in text token updates occurs specifically at Layer 1. This is followed by a prolonged, stable fusion phase (from Layers 4--20), during which both modalities exhibit low and consistent representation changes. In the final layers (21--28), a significant escalation is observed, resulting in sharp increases prior to the final output, with text token updates peaking at higher magnitudes than vision token updates.}
  \label{fig:rsqwen}
\end{minipage}\hfill
\begin{minipage}{0.48\textwidth}
  \centering
  \includegraphics[width=\linewidth]{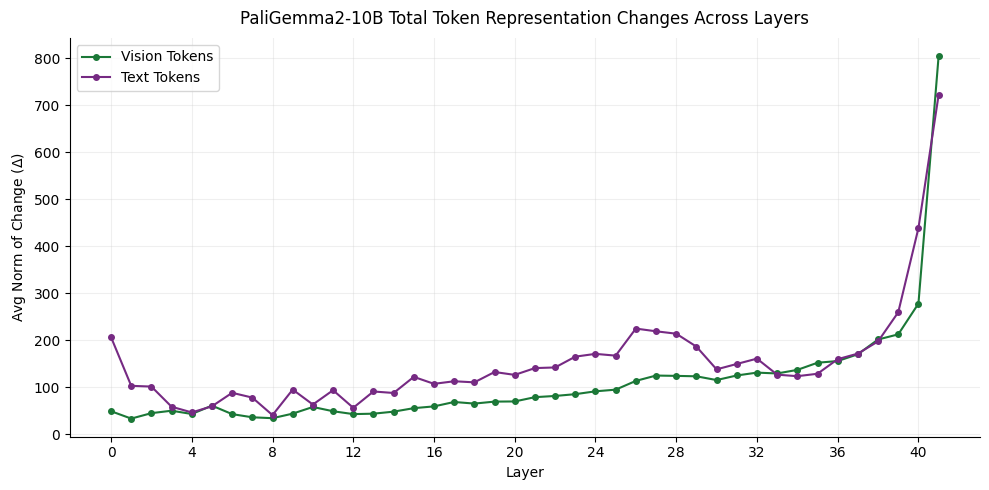}
  \Description{\textbf{Layer-wise residual stream update norms for PaliGemma2-10B.} The graph depicts the average norm of change ($\Delta$) across layers for vision and text tokens. The network architecture exhibits early modality asymmetry (Layers 0--24), where text tokens undergo larger, more volatile representation updates than their respective vision tokens. This transitions into a phase of deep mid-network alignment (Layers 25--36), characterized by converging update magnitudes. Finally, the model undergoes massive representation shifts in the layers immediately preceding unembedding, resulting in exponential spikes in which vision token updates ultimately exceed text token updates in absolute peak magnitude.}
  \caption{\textbf{Layer-wise residual stream update norms for PaliGemma2-10B.} The graph depicts the average norm of change ($\Delta$) across layers for vision and text tokens. The network architecture exhibits early modality asymmetry (Layers 0--24), where text tokens undergo larger, more volatile representation updates than their respective vision tokens. This transitions into a phase of deep mid-network alignment (Layers 25--36), characterized by converging update magnitudes. Finally, the model undergoes massive representation shifts in the layers immediately preceding unembedding, resulting in exponential spikes in which vision token updates ultimately exceed text token updates in absolute peak magnitude.}
  \label{fig:rsgemma10b}
\end{minipage}
\end{figure}

\subsection{Qwen2-VL}
\textsc{Qwen2-VL} reveals a processing strategy with foundational similarities to \textsc{PaliGemma-3B}, albeit on a different layer trajectory encompassing three primary phases:

\textbf{Layers 0-3:} While Layer 0 shows relatively low baseline updates, Layer 1 exhibits a massive spike for text tokens (magnitude $\approx 130$), while vision tokens remain entirely linear. This indicates \textsc{Qwen2-VL} establishes its text-specific representational shifts immediately, rather than in the mid-layers. By Layer 3, this spike resolves. 

\textbf{Layers 4-20:} After the early text bottleneck, the model enters a stable phase. Update trajectories for both modalities flatline at a very low magnitude, indicating a long phase of cross-modal fusion where visual features align with textual context without altering token identities.

\textbf{Layers 21-28:} Beginning around Layer 21, update magnitudes for both modalities drift upward. At Layer 28, text tokens reach $\approx 140$ and vision tokens reach $\approx 110$. Consistent with \textsc{PaliGemma-3B}, this final escalation indicates heavy transformation of the residual stream to align with the unembedding matrix for next-token prediction.

\subsection{PaliGemma2-10B}
We track \textsc{PaliGemma2-10B} to evaluate representation dynamics at scale. The model is characterized by early modality asymmetry followed by deep mid-network alignment:

\textbf{Layers 0-24:} The network begins with a high-magnitude projection for text tokens at Layer 0 ($\approx 200$), compared to a lower projection for vision embeddings ($\approx 50$). Throughout the first half of the network, a consistent modality gap is maintained. Text tokens gradually trend upward, whereas vision tokens remain exceptionally stable with low-magnitude updates, suggesting these early layers are dedicated to processing textual semantic context while preserving initial visual features.

\textbf{Layers 25-36:} Around Layer 25, the update magnitude for vision tokens rises to meet the text tokens. The trajectories converge and align closely, indicating a deep cross-modal fusion phase where equal computational effort is applied to both modalities to build a unified representation.

\textbf{Layers 37-41:} In the final layers, the fusion phase exponentially escalates as the model prepares the residual stream for the unembedding matrix. In the final layer (Layer 41), the magnitude of updates to vision tokens actually exceeds that applied to text tokens. This implies that at a 10B scale, projecting visual context into the text-based logit space requires a greater transformation than the text tokens themselves.

\section{High token semantic density influences the formation of the modality gap}\label{sec:appendix-high-token-semantic}
\subsection{Qwen2-VL-2B}
\begin{figure}[htbp]
    \centering
    \includegraphics[width=\linewidth]{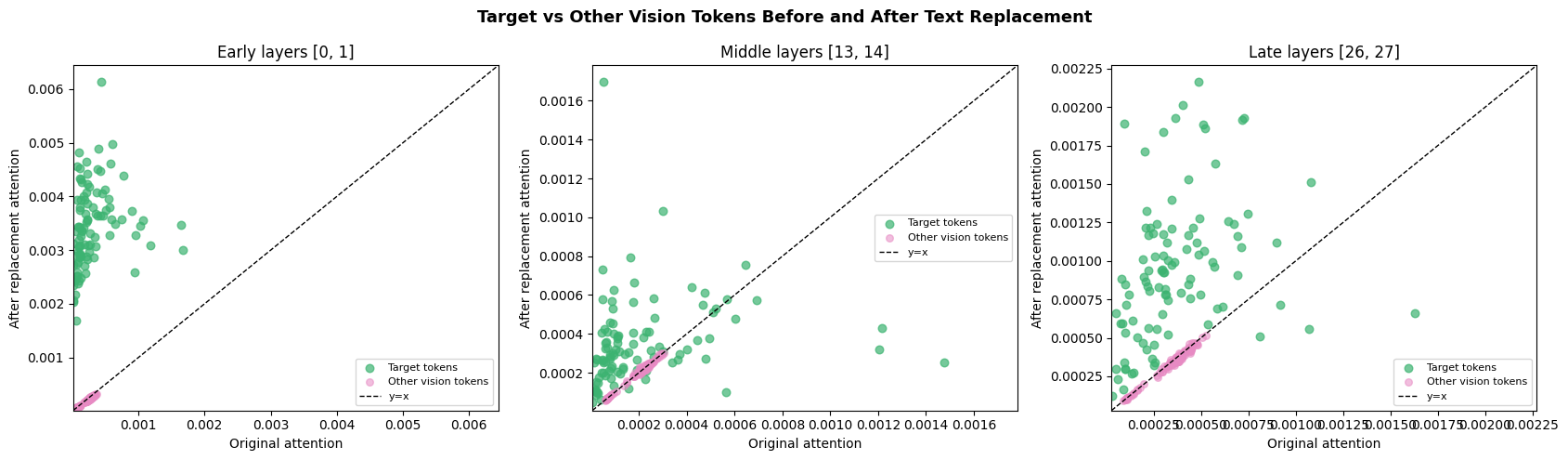}
    \Description{Traced 8 target tokens before and after text token replacement for \textsc{PaliGemma-3B}: Each dot represents a sample out of the 100 samples run. In all three graphs, attention is higher after the vision token target position is replaced with a text token, but the change varies. In early layers, attention to replaced positions is mostly condensed between 0.002 to 0.005. In intermediate layers, attention to replaced positions mostly stays under 0.001. In late layers, the attention to replaced tokens is higher with values ranging between 0.0005 to 0.002.}
    \caption{Traced 8 target tokens before and after text token replacement for \textsc{PaliGemma-3B}: Each dot represents a sample out of the 100 samples run. In all three graphs, attention is higher after the vision token target position is replaced with a text token, but the change varies. In early layers, attention to replaced positions is mostly condensed between 0.002 to 0.005. In intermediate layers, attention to replaced positions mostly stays under 0.001. In late layers, the attention to replaced tokens is higher with values ranging between 0.0005 to 0.002.}
    \label{fig:qwen-vision-token-replacement}
\end{figure} 
Replacing only the target vision tokens increases attention to those positions across all layer groups, with a strong effects in early layers. At late layers, this attention peaks. However, unreplaced vision tokens remain relatively stable. Similar to \textsc{PaliGemma-3B}, attention to other vision tokens only slightly decreases in early and late layers, while remaining fixed in middle layers. This indicates that semantic replacement selectively redirects attention toward the matched target positions without broadly disrupting the rest of the visual tokens.

\subsection{PaliGemma2-10B}

\begin{figure}[htbp]
    \centering
    \includegraphics[width=\linewidth]{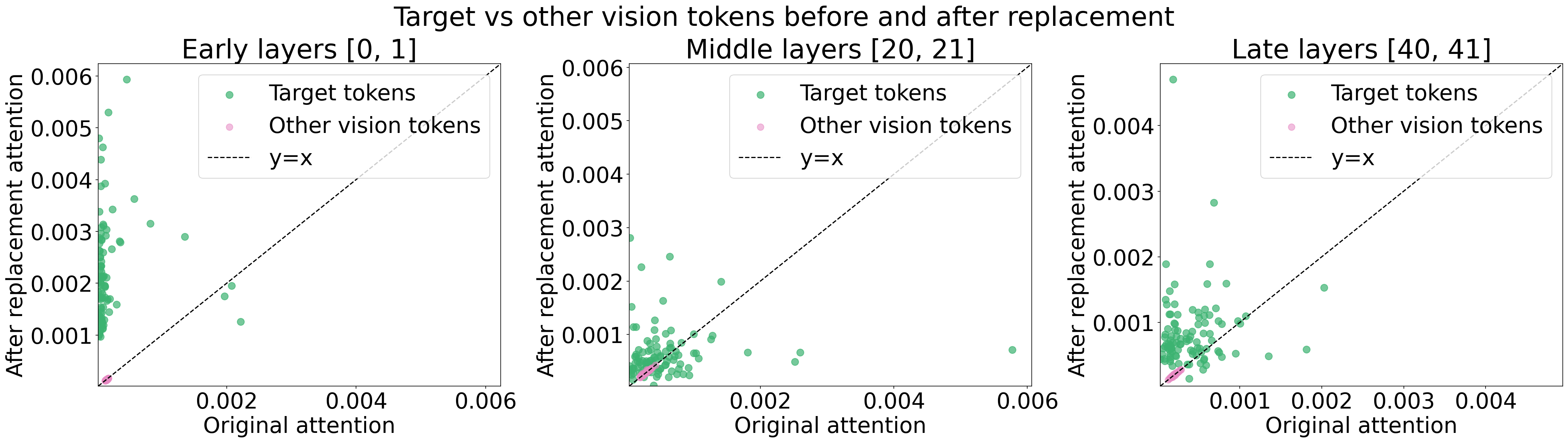}
    \Description{Traced 8 target tokens before and after text token replacement for \textsc{PaliGemma-3B}: Each dot represents a sample out of the 100 samples run. In all three graphs, attention is higher after the vision token target position is replaced with a text token, but the change varies. In early layers, attention to replaced positions is mostly condensed between 0.0005 to 0015. In intermediate layers, attention to replaced positions mostly stays under 0.001. In late layers, the attention to replaced tokens is much higher with values ranging between 0.002 to 0.004.}
    \caption{Traced 8 target tokens before and after text token replacement for \textsc{PaliGemma-3B}: Each dot represents a sample out of the 100 samples run. In all three graphs, attention is higher after the vision token target position is replaced with a text token, but the change varies. In early layers, attention to replaced positions is mostly condensed between 0.0005 to 0015. In intermediate layers, attention to replaced positions mostly stays under 0.001. In late layers, the attention to replaced tokens is much higher with values ranging between 0.002 to 0.004.}
    \label{fig:paligemma2-vision-token-replacement}
\end{figure} 

Replacing only the target vision tokens increases attention to those positions across all layer groups, with the strongest effect in early layers. However, unreplaced vision tokens remain mostly stable across depth. Attention to other vision tokens changes minimally in early, middle, and late layers. However, rather than experiencing a decrease in attention like \textsc{PaliGemma-3B} and \textsc{Qwen2-VL}, the values increase at middle and late layers. While it leaves the attention at other vision tokens largely unchanged, attention seems to increase overall for both the replaced target and unchanged vision tokens.

\section{Formal Metric Definitions}
\label{sec:metric-definitions}
We formally define the metrics used to evaluate the models' cross-modal behavior. Let $L$ be the number of transformer layers; $H^{(l)} \in \mathbb{R}^{T \times d}$ denote the hidden state activations at layer $l$, where $T$ is the sequence length and $d$ is the hidden dimension; and $I_{\text{vis}}$ and $I_{\text{text}}$ denote the sets of indices corresponding to vision and text tokens, respectively.

\subsection{Cosine Similarity of Representations}\label{sec:cosine-metrics}
To measure the stability of visual representations across consecutive layers, we compute the cosine similarity between the hidden state of a token at layer $l$ and layer $l+1$. For a specific vision token at position $t \in I_{\text{vis}}$, the similarity is defined as:
\begin{equation}
\text{CosSim}(t, l) = \frac{H_t^{(l)} \cdot H_t^{(l+1)}}{\|H_t^{(l)}\|_2 \|H_t^{(l+1)}\|_2}
\end{equation}
The reported metric is the average similarity over all vision tokens across $N$ samples in the PixelProse dataset:
\begin{equation}
\text{AvgSim}^{(l)} = \frac{1}{N} \sum_{k=1}^{N} \left( \frac{1}{|I_{\text{vis}}|} \sum_{t \in I_{\text{vis}}} \text{CosSim}_k(t, l) \right)
\end{equation}

\subsection{Residual Stream Update \& Decomposition}
The transformer residual stream at layer $l$ is updated by two distinct sub-modules: Multi-Head Attention (Attn) and the Multi-Layer Perceptron (MLP). Within the transformer stack of an adapter-based VLM (assuming a standard pre-LN configuration), the hidden state update is defined as:
\begin{equation}
H^{(l)} = H^{(l-1)} + \Delta_{\text{Attn}}^{(l)} + \Delta_{\text{MLP}}^{(l)}
\end{equation}
where the individual contributions are:
\begin{align}
\Delta_{\text{Attn}}^{(l)} &= \text{Attn}(\text{LN}(H^{(l-1)})) \\
\Delta_{\text{MLP}}^{(l)} &= \text{MLP}(\text{LN}(H^{(l-1)} + \Delta_{\text{Attn}}^{(l)}))
\end{align}
The \textbf{Residual Stream Update} metric quantifies the magnitude of change introduced by layer $l$. We calculate the $L_2$ norm of the total difference between the pre-layer and post-layer representations for a specific modality set $S \in \{I_{\text{vis}}, I_{\text{text}}\}$:
\begin{equation}
\text{Update}^{(l)}_S = \frac{1}{|S|} \sum_{t \in S} \| H_t^{(l)} - H_t^{(l-1)} \|_2
\end{equation}
To determine the drivers of this processing, we calculate the \textbf{Relative Contribution} of each sub-module by averaging the per-token norms:
\begin{equation}
\text{Contrib}_{\text{Attn}}^{(l)} = \frac{1}{|S|} \sum_{t \in S} \frac{\|\Delta_{\text{Attn}, t}^{(l)}\|_2}{\|\Delta_{\text{Attn}, t}^{(l)}\|_2 + \|\Delta_{\text{MLP}, t}^{(l)}\|_2}
\end{equation}

\subsection{Attention by Modality}
We analyze the attention distribution of the final token (the query $q_{last}$), which is responsible for generating the next token. Let $A^{(l)}_{last, j}$ represent the attention weight from the last token to key token $j$ at layer $l$, averaged across all attention heads $h$. Assuming $Q$ and $K$ denote the Query and Key projections respectively:
\begin{equation}
A^{(l)}_{last, j} = \frac{1}{N_{heads}} \sum_{h=1}^{N_{heads}} \text{Softmax}\left(\frac{Q_{last}^{(l,h)} (K_j^{(l,h)})^T}{\sqrt{d_k}}\right)
\end{equation}
We define the \textbf{Modality Attention Score} as the total probability mass allocated to a specific modality (Vision or Text):
\begin{equation}
\text{Attn}_{\text{vision}}^{(l)} = \sum_{j \in I_{\text{vis}}} A^{(l)}_{last, j}, \quad \quad \text{Attn}_{\text{text}}^{(l)} = \sum_{j \in I_{\text{text}}} A^{(l)}_{last, j}
\end{equation}

\subsection{Averaging Procedures}
All reported results are averaged over the PixelProse dataset. For any metric $M$ computed for a single sample $k$, the final reported value is:
\begin{equation}
\bar{M} = \mathbb{E}_{x \sim \mathcal{D}} [M(x)] \approx \frac{1}{N} \sum_{k=1}^{N} M_k
\end{equation}
Where $N=100$ for our representational analysis experiments.

\clearpage

\begin{center}
    \includegraphics[width=0.32\linewidth]{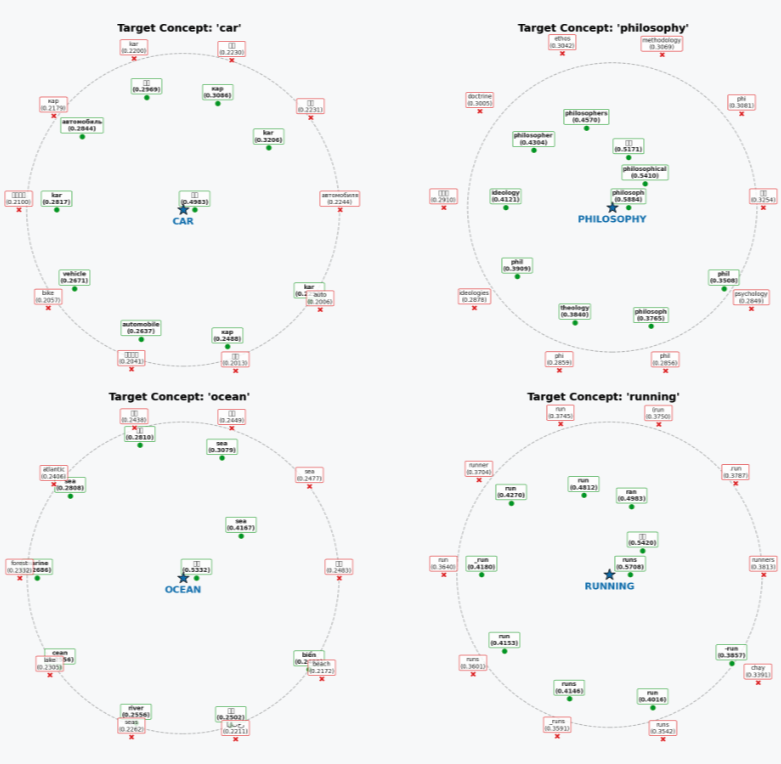}\hfill
    \includegraphics[width=0.32\linewidth]{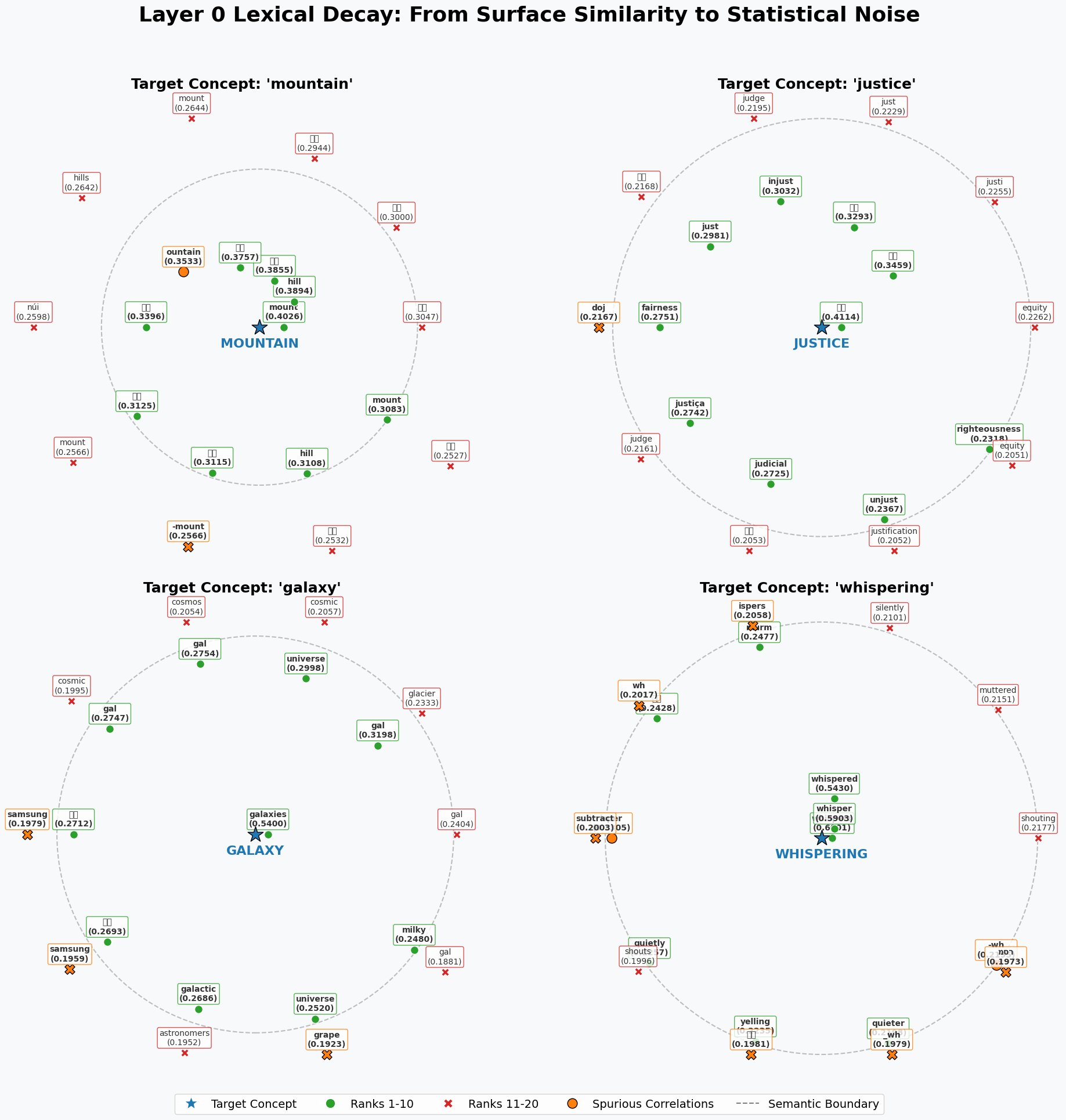}\hfill
    \includegraphics[width=0.32\linewidth]{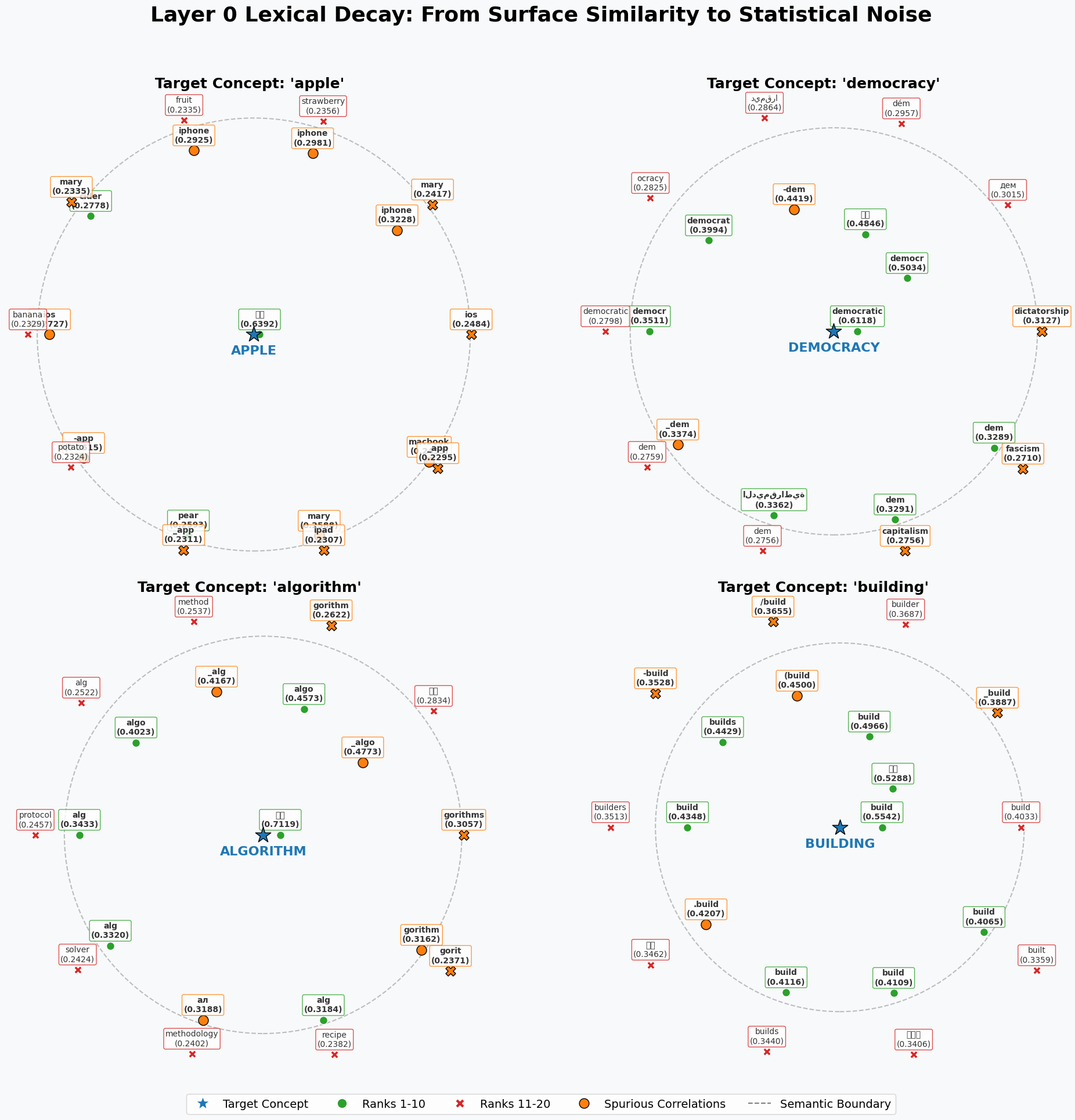}
\end{center}
\captionof{figure}{\textbf{\textsc{Qwen2-VL-2B} text kNN examples.} \textit{Left:} Four target concepts (`car', `philosophy', `ocean', and `running') mapped against their top-20 nearest neighbors by cosine similarity. \textit{Middle:} Four additional concepts (`mountain', `justice', `galaxy', and `whispering') showing rapid decay past the semantic boundary into statistical noise, including tech-brand contamination for ``galaxy'' (``samsung''), institutional acronym bias for ``justice'' (``doj''), and sub-word token fragmentation for ``whispering'' (``subtract'' and ``-wh''). \textit{Right:} Four further concepts (`apple', `democracy', `algorithm', and `building') illustrating tech-ecosystem contamination for ``apple'' (``iphone'' and ``macbook''), competing political ideologies for ``democracy'' (``dictatorship'' and ``fascism''), and code syntax fragmentation (``\_algo'' and ``/build'').}
\label{fig:qwen-text-knn-all}

\bigskip

\begin{center}
    \includegraphics[width=0.32\linewidth]{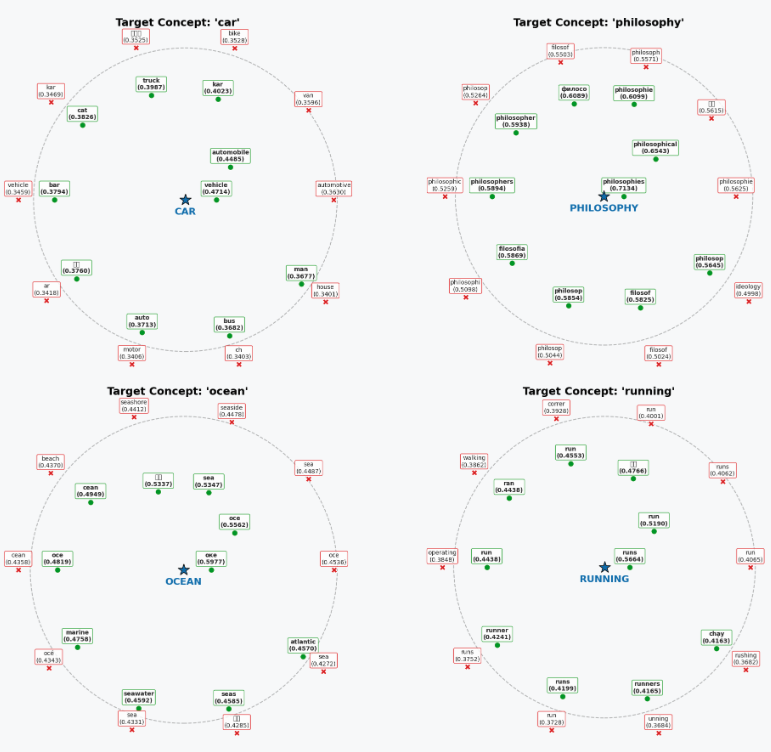}\hfill
    \includegraphics[width=0.32\linewidth]{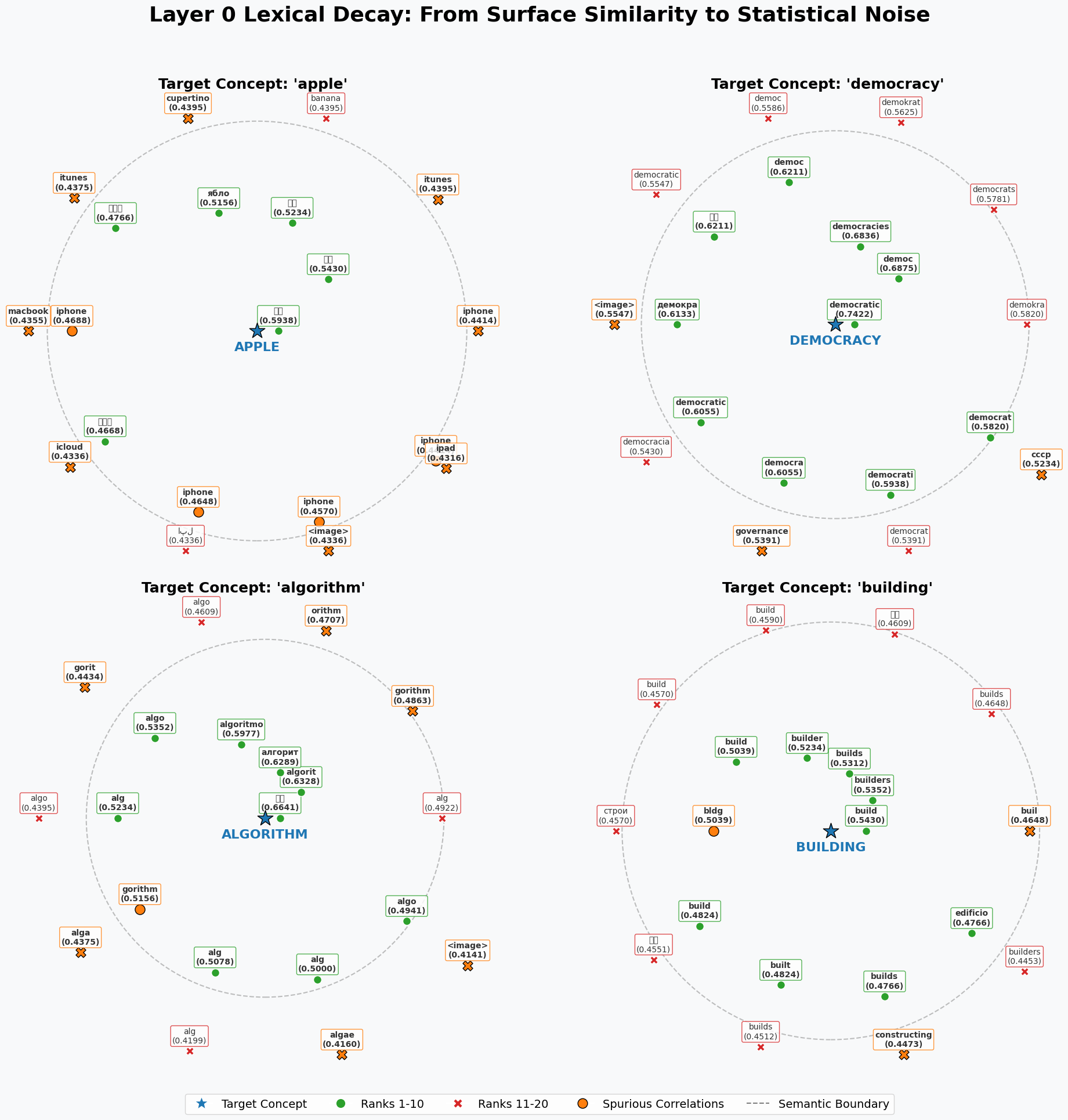}\hfill
    \includegraphics[width=0.32\linewidth]{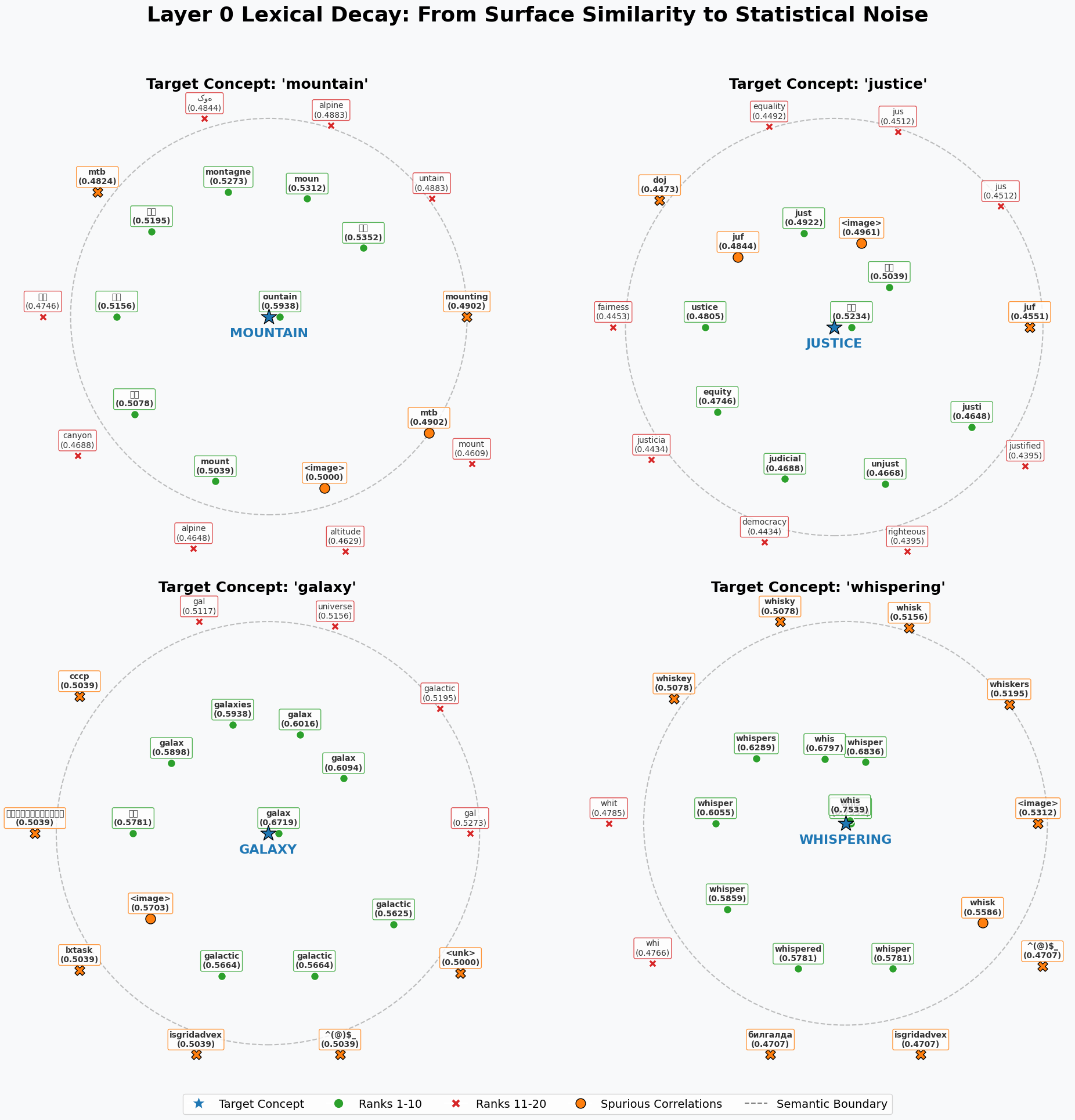}
\end{center}
\captionof{figure}{\textbf{\textsc{PaliGemma2-10B} text kNN examples.} \textit{Left:} Four target concepts (`car', `philosophy', `ocean', and `running') mapped against their top-20 nearest neighbors by cosine similarity. \textit{Middle:} Four additional concepts (`apple', `democracy', `algorithm', and `building'): high lexical similarity in ranks 1--10 quickly decays past the semantic boundary into statistical noise, exposing pre-training biases such as tech-ecosystem contamination for ``apple'' (``iphone'' and ``macbook'') and structural fragmentation for ``algorithm'' (``algae'' and ``gorit''). \textit{Right:} Four further concepts (`mountain', `justice', `galaxy', and `whispering'): high lexical similarity in ranks 1--10 again decays into noise, including severe lexical contamination for ``whispering'' (``whiskey'' and ``whiskers'') and raw system variables/code syntax for ``galaxy'' (``isgridadvex'' and \texttt{\^{}(@)\$\_}).}
\label{fig:paligemma2-text-knn-all}

\newpage

\begin{center}
    \includegraphics[width=\linewidth]{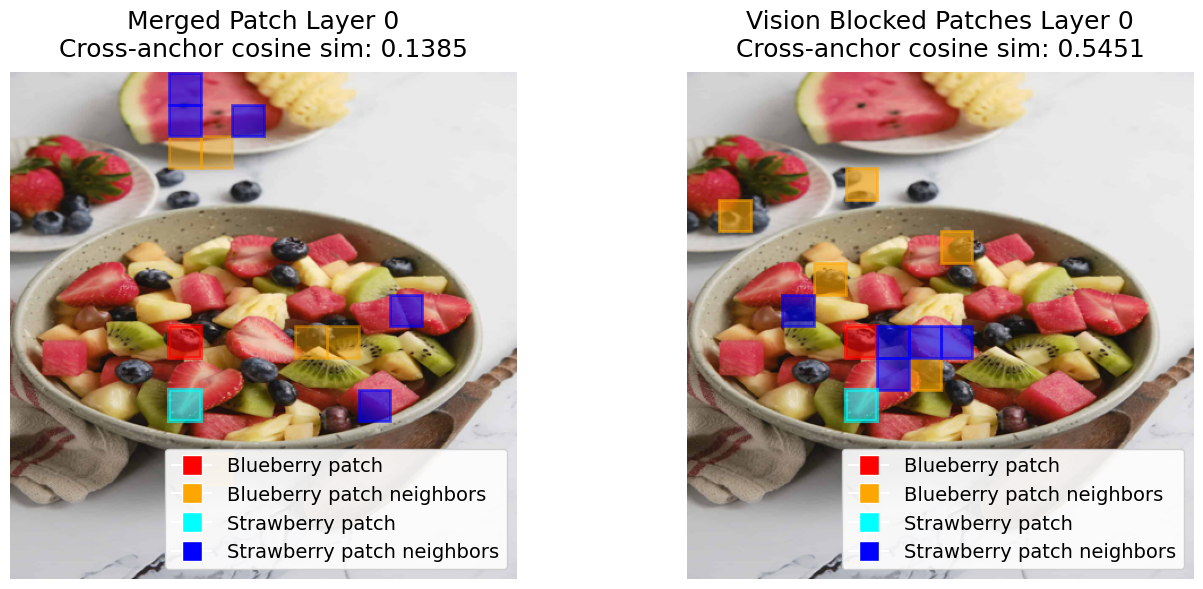}\\[4pt]
    \includegraphics[width=\linewidth]{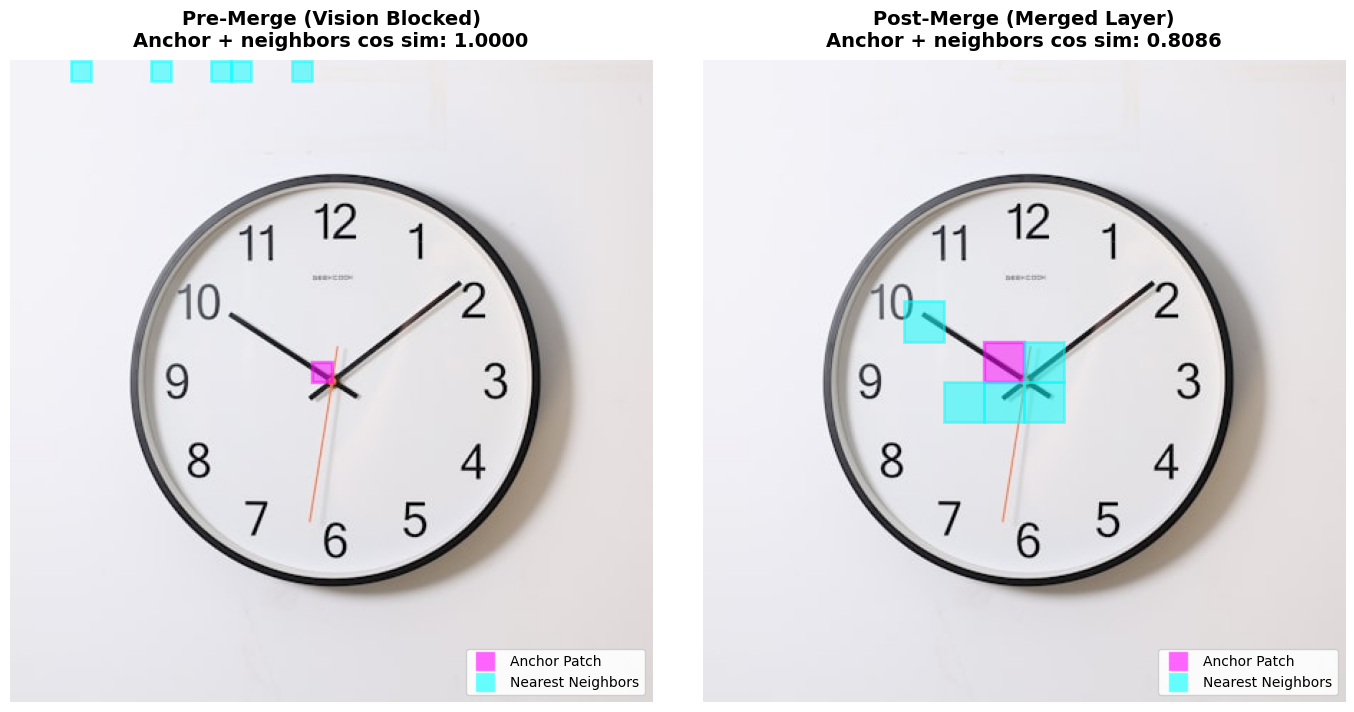}\\[4pt]
    \includegraphics[width=\linewidth]{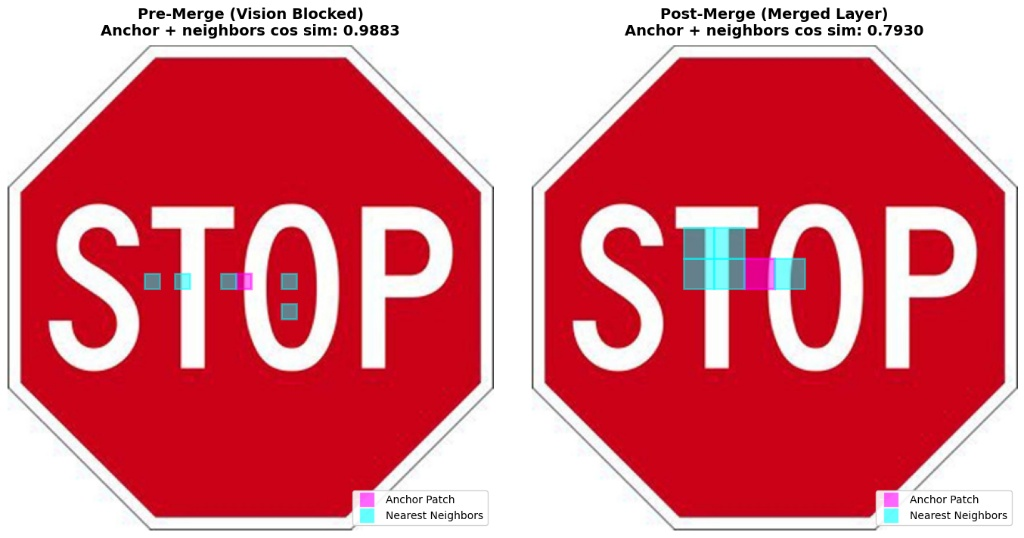}
\end{center}
\captionof{figure}{\textbf{\textsc{Qwen2-VL-2B} image kNN examples.} \textit{Top:} Semantic density within vision tokens: in the initial merged patch layer, the strawberry patch neighbors are concentrated on red patches, while the blueberry patch neighbors show patches that are half-dark and half-light. After vision-blocking, the vision model properly identifies neighbors that we would normally associate with the target token. \textit{Middle:} An anchor on the clock's center axis in the Pre-Merge layer incorrectly retrieves uniform background patches at the top edge. Following the Patch Merger, true semantic density is established, correctly clustering nearest neighbors around the clock hands. \textit{Bottom:} The anchor token placed at the center of the Stop Sign has neighbors randomly concentrated in the Pre-Merge layer. The Post-Merge layer tightly clusters the nearest neighbors exclusively on the white text characters of ``STOP'', demonstrating cross-modal text alignment.}
\label{fig:qwen-vision-knn-all}

\newpage

\begin{center}
    \includegraphics[width=\linewidth]{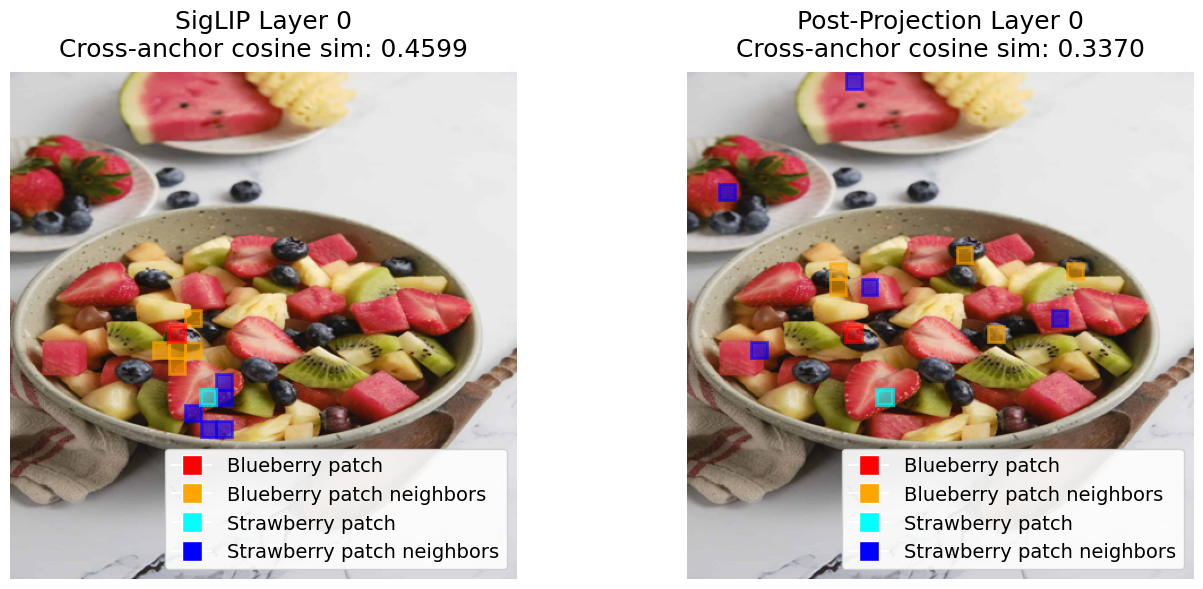}\\[4pt]
    \includegraphics[width=\linewidth]{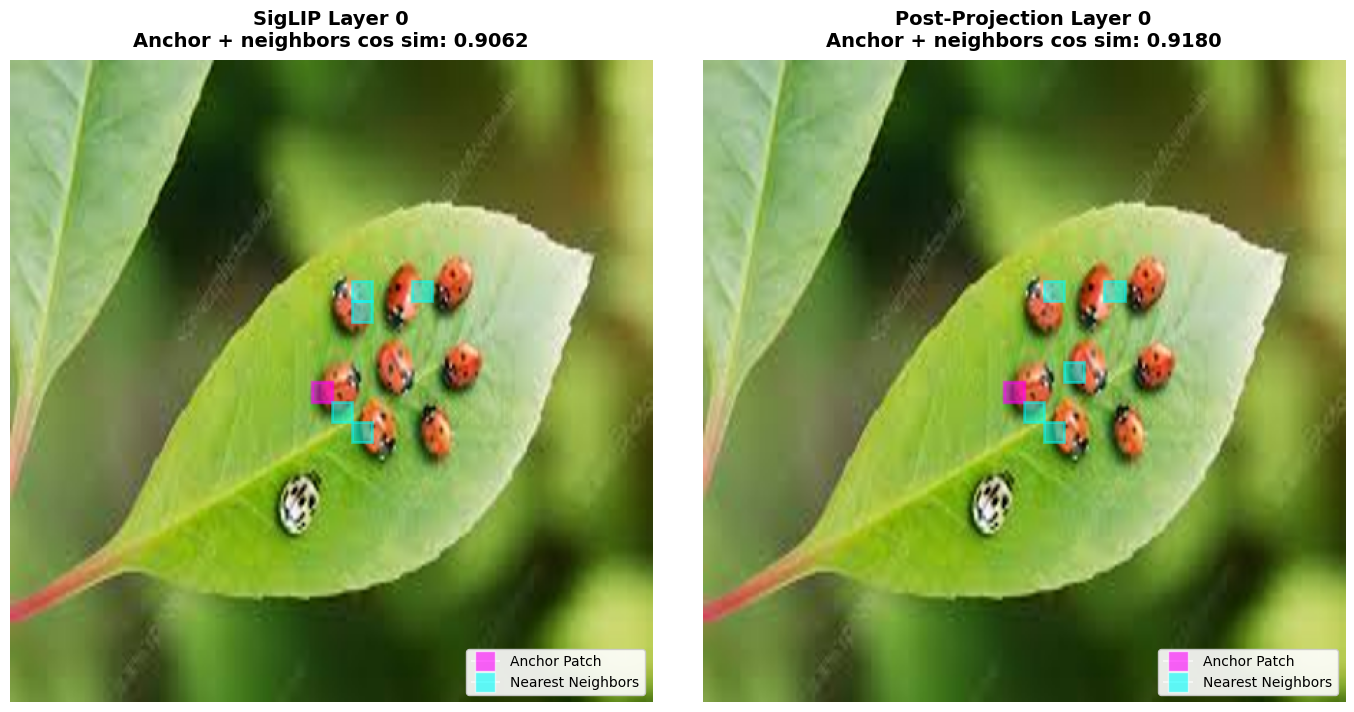}\\[4pt]
    \includegraphics[width=\linewidth]{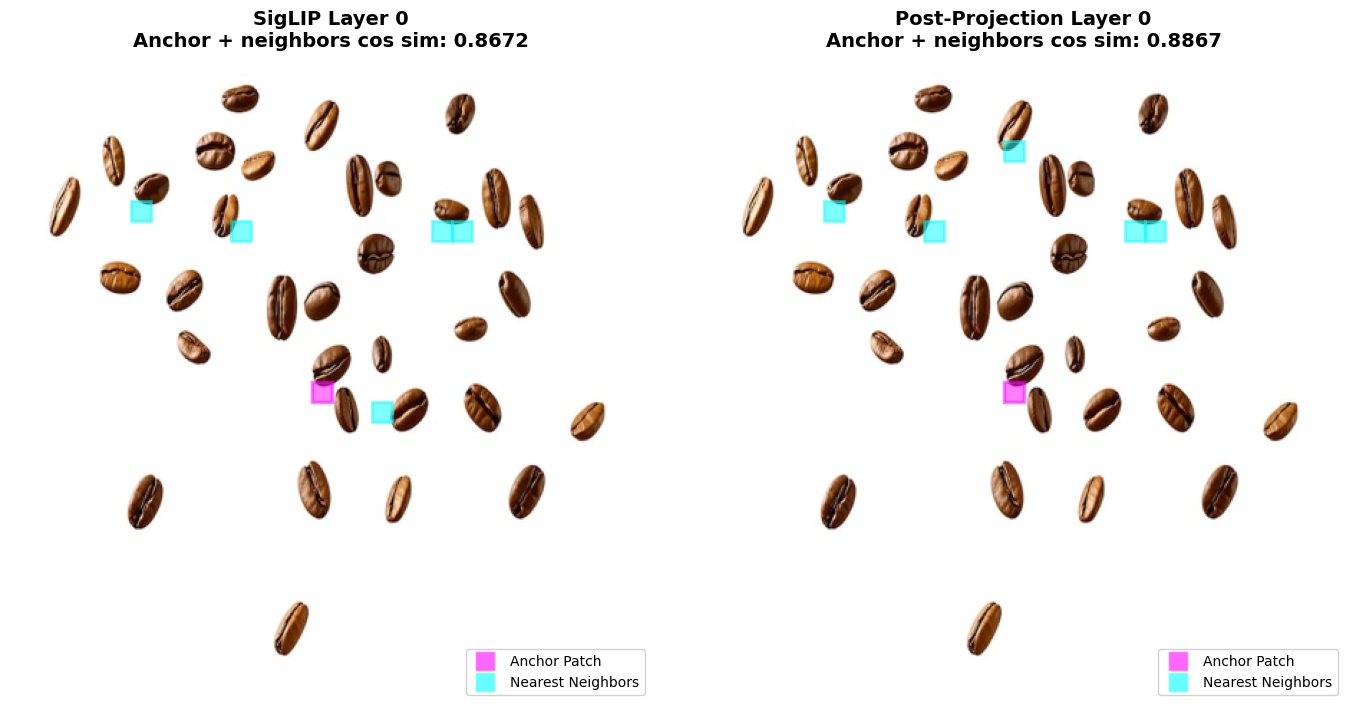}
\end{center}
\captionof{figure}{\textbf{\textsc{PaliGemma2-10B} image kNN examples.} \textit{Top:} Semantic density within vision tokens: in the initial layer, both the strawberry patch neighbors and the blueberry patch neighbors are nearby the initial target token. Post-projection, the vision model properly identifies neighbors that we would normally associate with the target token. \textit{Middle:} In the pre-projection layer, an anchor placed on a single red ladybug retrieves matching semantic neighbors while successfully ignoring the leaf background and the differing white ladybug. Post-projection, the vision model properly preserves this spatial and semantic clustering, thereby maintaining associations with the target token's neighbors. \textit{Bottom:} Initially, an anchor placed on a single scattered coffee bean retrieves tokens corresponding to other individual beans. Post-projection, the vision model preserves this spatial and semantic clustering, highlighting the linear adapter's geometry-preserving nature.}
\label{fig:paligemma2-vision-knn-all}

\end{document}